\documentclass{article}
\usepackage{ijcai21}

\usepackage{times}

\usepackage{soul}
\usepackage{url}
\usepackage[hidelinks]{hyperref}
\usepackage[utf8]{inputenc}
\usepackage[small]{caption}
\usepackage{graphicx}
\usepackage{amsmath}
\usepackage{booktabs}
\usepackage{booktabs}
\usepackage{amsmath}
\usepackage{xcolor}
\usepackage{amsmath,amssymb,amsthm}

\usepackage{url}

\usepackage{subfigure}
\usepackage{multirow}

\newcommand{\mycite}[1]{\citeauthor{#1}~\shortcite{#1}}

\usepackage{xcolor}

\title{Modelling General Properties of Nouns\\ by Selectively Averaging Contextualised Embeddings}

\author{
Na Li$^1$\footnote{Contact Author}\and
Zied Bouraoui$^2$\and
Jose Camacho-Collados$^3$ \and \\ 
Luis Espinosa-Anke$^3$\and
Qing Gu$^1$\and
Steven Schockaert$^3$\\
\affiliations
$^1$Nanjing University, China; 
$^2$CRIL CNRS \& Univ Artois, France;
$^3$Cardiff University, UK\\
\emails
li.na@smail.nju.edu.cn,
zied.bouraoui@cril.fr,
\{camachocolladosj, espinosa-ankel,schockaerts1\}@cardiff.ac.uk,
guq@nju.edu.cn
}

\begin{document}

\maketitle

\begin{abstract}
While the success of pre-trained language models has largely eliminated the need for high-quality static word vectors in many NLP applications, such vectors continue to play an important role in tasks where words need to be modelled in the absence of linguistic context.
In this paper, we explore how the contextualised embeddings predicted by BERT can be used to produce high-quality word vectors for such domains, in particular related to knowledge base completion, where our focus is on capturing the semantic properties of nouns. We find that a simple strategy of averaging the contextualised embeddings of masked word mentions leads to vectors that outperform the static word vectors learned by BERT, as well as those from standard word embedding models, in property induction tasks. We notice in particular that masking target words is critical to achieve this strong performance, as the resulting vectors focus less on idiosyncratic properties and more on general semantic properties. Inspired by this view, we propose a filtering strategy which is aimed at removing the most idiosyncratic mention vectors, allowing us to obtain further performance gains in property induction. 
\end{abstract}

%%%%%%%%%%%%%%%%%%%%%%%%%%%%%%%%
\section{Introduction}
The success of contextualised language models (LMs), such as BERT \cite{DBLP:conf/naacl/DevlinCLT19}, has led to a paradigm shift in Natural Language Processing (NLP). %LMs capture prior knowledge about word meaning, and language more generally, in a powerful but opaque way. 
A key feature of such models is that they produce contextualised word vectors, i.e.\ vectors that represent the meaning of words in the context of a particular sentence. While the shift away from standard word embeddings has benefited an impressively wide range of NLP tasks, many other tasks still crucially rely on static representations of word meaning. For instance, in information retrieval, query terms often need to be modelled without any other context, and word vectors are commonly used for this purpose \cite{onal2018neural}. %In entity retrieval, using word vectors is particularly crucial to match query terms to the vector encodings of candidate entities \cite{nikolaev2020joint}. 
In zero shot learning, word vectors are used to obtain category embeddings \cite{socher2013zero,ma2016label}. Word vectors are also used in topic models \cite{das2015gaussian}. In the context of the Semantic Web, word vectors have been used for ontology alignment \cite{kolyvakis2018deepalignment}, concept invention \cite{vimercati2019mapping} and ontology completion \cite{li2019ontology}. 
%\red{In this paper we focus on this latter type of application XXX COMMENT: not sure if this is entirely true, but would be nice to have a link with the paper, as otherwise it reads as just a list of applications without much connection.}
%
The word vectors learned by standard word embedding models, such as Skip-gram \cite{DBLP:journals/corr/abs-1301-3781} and GloVe \cite{DBLP:conf/emnlp/PenningtonSM14}, essentially summarise the contexts in which each word occurs. However, these contexts are modelled in a shallow way, capturing only the number of co-occurrences between target words and individual context words. The question we address in this paper is whether the more sophisticated context encodings that are produced by LMs can be used to obtain higher-quality word vectors. Motivated by the aforementioned applications, we focus in particular embeddings that capture the semantic properties of nouns.

%We focus in particular on modelling nouns, as this is what is primarily needed in the aforementioned applications. 
To obtain static word vectors from a contextualised language model, we can sample sentences in which a given word $w$ occurs, obtain a contextualised vector representation of $w$ from these sentences, and finally average these vectors \cite{DBLP:conf/acl/BommasaniDC20,DBLP:conf/emnlp/VulicPLGK20}. In this paper, we aim to improve on this strategy. First, when obtaining a contextualised representation of the target word, we replace that target word by the [MASK] token. This has the advantage that words which consist of multiple word-pieces can be modelled in a natural way. More fundamentally, our hypothesis is that this will lead to representations that are more focused on the semantic properties of the given word. In particular, because the target word is masked, the resulting vector specifically captures what that sentence reveals about the word. A bag of masked sentences can thus be viewed as a bag of properties.
%
%To see how LMs can be used to learn static word vectors, consider a sentence in which some noun $w$ appears. We can mask the occurrence of $w$ and then use a masked language model such as BERT to predict the masked word. This prediction takes the form of a vector $\mathbf{m}$, which implicitly encodes a probability distribution over words. We will refer to $\mathbf{m}$ as a mention vector. Intuitively, this vector encodes what the given sentence reveals about the nature of $w$. In other words, we can view $\mathbf{m}$ as representing a property which is satisfied by $w$. By computing mention vectors $\mathbf{m_1},...,\mathbf{m_n}$ for all occurrences of $w$ in the corpus, we thus obtain a description of $w$ as a bag of properties. 
%
To illustrate this, consider the Wikipedia sentences in Table \ref{TabBananas}, where occurrences of the word \emph{bananas} were masked. From BERT's top predictions for the missing word, we can see that these sentences indeed reveal different properties of bananas, e.g.\ being edible, a dessert ingredient and a type of fruit.  %\todo{Another strategy would be to compute mention vectors from sentences without masked LM....}

\begin{table}
\footnotesize
\begin{tabular}{p{105pt}p{107pt}}
\toprule
\textbf{Masked sentence} & \textbf{BERT predictions}\\
%\midrule
%\emph{The teacher then hits upon the fact he has yet to teach them about \underline{\phantom{aaa}}.} &  it, love, him, sex, science, religion, magic, this, music, god\\
\midrule
\emph{\underline{\phantom{aaa}} are cultivated by both small farmers and large land holders.} & they, these, crops, fields, potatoes, gardens, vegetables, most, vines, trees\\
\midrule
\emph{Banoffee pie is an English dessert pie made from \underline{\phantom{aaa}}, cream and toffee ...} & cheese, sugar, butter, apples, eggs, milk,  chocolate, 
honey, apple, egg\\
\midrule
\emph{\underline{\phantom{aaa}} are a popular fruit consumed worldwide with a yearly production of over ...} &  they, bananas, citrus, apples, grapes, these, fruits, potatoes, berries, nuts\\
\bottomrule
\end{tabular}
\caption{Top predictions from BERT-large-uncased for sentences.\label{TabBananas}}
\end{table}

This view of masked sentences as encoding properties suggests another improvement over plain averaging of contextualised vectors. Since some properties are intuitively more important than others, we should be able to improve the final representations by averaging only a particular selection of the contextualised vectors.  Consider the following Wikipedia sentence: \emph{Banana equivalent dose (BED) is an informal measurement of ionizing radiation exposure}.
By masking the word ``banana'', we can obtain a contextualised vector, but this vector would not capture any of the properties that we would normally associate with bananas.  The crucial difference with the sentences from Table \ref{TabBananas} is that the latter capture \emph{general properties}, i.e.\ properties that apply to more than one concept, whereas the sentence above captures an \emph{idiosyncratic property}, i.e.\ a property that only applies to a particular word.
%Broadly speaking, we can distinguish between those that capture idiosyncratic properties \steven{(i.e.\ properties that only apply to a particular noun)} and those that capture more general properties. 
In the aforementioned application domains, static word vectors are essentially used to capture the commonalities between given sets of words (e.g.\ between training and test categories in zero shot learning). %Essentially, we then need to identify the semantic properties that distinguish a given a set words from others. 
Word vectors should thus primarily capture the general properties of nouns. 
Inspired by this view, we propose a simple strategy for identifying contextualised vectors that are likely to capture idiosyncratic properties. When computing the vector representation of a given noun, we then simply omit these mention vectors, and compute the average of the remaining ones.

%\smallskip
%\noindent The main contributions of this paper are as follows:
%\begin{enumerate}
%\item We analyse the potential of averaged mention vectors as static word vectors, finding that they are particularly suitable for modelling semantic properties, although we also find that they underperform standard word embeddings when it comes to modelling word similarity.
%\item We propose and evaluate a strategy which is aimed at removing mention vectors that capture idiosyncratic properties. We find that this strategy consistently improves the quality of the resulting averaged mention vectors.
%\item As an example of a downstream application that critically relies on static word representations, we empirically show the usefulness of our strategy for ontology completion.
%\end{enumerate}

%*******************************
\section{Related Work}

Several authors have analysed the extent to which pre-trained LMs capture semantic knowledge. For instance, \mycite{forbes2019neural} focused on predicting the properties of concepts (i.e.\ nouns), as well as their affordances (i.e.\ how they can be used). %They also looked at whether LMs can predict the affordances of concepts from their properties, and vice versa, which proved to be a particularly challenging task. 
\mycite{weir2020existence} considered the following problem: given a sentence that specifies the properties of a given concept, can LMs be used to predict the concept? For instance, given the input \emph{``A [MASK] is tasty, is eaten, is made of sugar, is made of flour, and is made of eggs''}, the aim is to predict \emph{cake}. %Somewhat related, \mycite{ettinger2020bert} analyses whether BERT's predictions are sensitive to negation and semantic role, and whether BERT is able to make predictions that require context from a preceding sentence. Apart from modelling the properties of concepts, some work has also looked at relational knowledge. For instance, \mycite{langmodelknowledgebases2019} showed that pre-trained LMs can to some extent be used for link prediction, by converting queries into masked sentences, using manually encoded templates (e.g.\ \emph{``The official language of Mauritius is [MASK]''}). A similar strategy is used by \mycite{davison2019commonsense} for commonsense knowledge graph completion. Some work has also looked at methods for constructing such templates automatically \cite{bouraoui2020inducing,jiang2020can}. Another strategy, explored by \mycite{bosselut-etal-2019-comet}, is to fine-tune the LM such that it can take direct encodings of knowledge graph triples as input. 
%The aforementioned works do not focus on learning word representations, using LMs instead in a more direct fashion.
Some previous work has already explored the idea of using the contextualised word vectors predicted by neural LMs for modelling word meaning. For instance, \mycite{amrami2019towards} obtain the set of contextualised vectors predicted by BERT, for different mentions of the same word, and then cluster these vectors to perform word sense induction. %Similar strategies have also been used for word sense disambiguation \cite{DBLP:conf/acl/LoureiroJ19,vial2019sensecompresswsd}.
\mycite{mickus2019you} analyse the distribution of contextualised word vectors, finding that the vectors corresponding to the same word type are largely clustered together. The distribution of contextualised word vectors is also studied by 
\mycite{DBLP:conf/emnlp/Ethayarajh19}, who furthermore explores the idea of learning static word vectors by taking the first principal component of the contextualised vectors of a given word (which has a similar effect as taking their average). They find that the vectors obtained from the earlier layers of BERT (and other LMs) perform better in word similarity tasks than the later layers. In contrast, \mycite{DBLP:conf/acl/BommasaniDC20} found that the optimal layer depends on the number of sampled mentions, with later layers performing better with a large number of mentions. Rather than fixing a single layer, \mycite{DBLP:conf/emnlp/VulicPLGK20} advocated averaging representations from several layers.  

%They also find that the distribution of contextualised vectors for a given word type is highly anisotropic. Importantly, the aforementioned works are based on contextualised vectors that are obtained without masking, which is why even early layers capture word meaning. As we will show, by masking the input word, we can obtain contextualised vectors that are better suited for modelling general semantic properties.

%Earlier, \mycite{erk2009representing} introduced the idea of learning one vector for each occurrence of a word, viewing word types as occupying a region in the vector space. A similar assumption is also used by embedding models that represent words as Gaussians \cite{vilnis2014word} or as quadratic surfaces \cite{DBLP:conf/conll/JameelS17}. 

%*******************************
\section{Encoding Words with Mention Vectors}
In this section we describe our strategy for obtaining static word vectors from BERT. First, we provide some background on the BERT model and we explain how we use BERT to obtain mention vectors for a given word. %In Section \ref{secAnalysis}, we then present some analysis of the differences between the static word vectors that are learned by the BERT model itself, and the averaged mention vectors that we compute. Finally, 
Finally, we also introduce our proposed filtering strategy.

\paragraph{Background and Notation}
%We rely on a pre-trained masked language model; in this paper we will use BERT for this purpose. 
BERT represents text fragments as sequences of tokens. Frequent words are represented as a single token, whereas less common words are represented as sequences of sub-word tokens, called word-pieces. Given an input sentence $S=t_1...t_k$, BERT predicts a contextualised vector $\phi_i(S)$ for each token $t_i$. Together with this mapping $\phi$, which takes the form of a deep transformer model, BERT learns a static vector $\mathbf{t}_{\textsc{static}}$ for each word-piece $t$ in its vocabulary $V_{\textsc{BERT}}$. During training, some tokens are replaced by a special token [MASK]. If this is the case for the i\textsuperscript{th} token of the sentence $S$, $\phi_i(S)$ encodes a probability distribution over word-pieces, corresponding to BERT's prediction of which token from the original sentence was masked. 
%It is defined as follows ($w\in V_{\textsc{BERT}}$):
%$$
%P(w;S,i) = \frac{\exp(\phi_i(S) \cdot \mathbf{w}_{\textsc{static}})}{\sum_{v\in V_{\textsc{BERT}}} \exp(\phi_i(S) \cdot \mathbf{v}_{\textsc{static}})}
%$$
%where we write $V_{\textsc{BERT}}$ for the set of word-pieces in the BERT vocabulary and $\mathbf{w}_{\textsc{static}}$ for the (static) vector representation of the word-piece $w$. % \todo{and $\mathbf{w}_{\textsc{static}}$ is  a (static) vector representation of a given word.}

\paragraph{Obtaining Mention Vectors}
Let $W$ be the set of nouns for which we want to learn a  vector representation. For each $w \in W$, we randomly sample $N$ mentions of $w$ from a given corpus. From each of the corresponding sentences, we obtain a vector by masking the occurrence of $w$ and taking the contextualised vector predicted by BERT for the position of this [MASK] token. We will refer to this vector as a mention vector. For $w\in W$, we write $\mu(w)$ for the set of all mention vectors that are obtained for $w$. 
%collect a bag of mention vectors $\mu(w) = \{\mathbf{m}_1,...,\mathbf{m}_{n_w}\}$ from a given text corpus. Each of these mention vectors is the prediction $\phi_i(S)$ for a sentence from the corpus in which $w$ occurs, when that occurrence of $w$ is masked. %We use the term mention vector to distinguish these vectors from the contextualised word vectors that can be obtained without masking words, e.g.\ as used for Word Sense Disambiguation. 
%The most straightforward way to learn a single vector representation of $w$ from the set of mention vectors is to average them:
%$$
%\mathbf{w}_{\textsc{Avg}} = \text{Avg}(\mu(w))
%$$
%where for a bag of vectors $X$ we define:
%$$
%\text{Avg}(X) =
%\frac{1}{|X|} \sum_{\mathbf{x} \in X} \mathbf{x}
%$$
A key design choice in our approach is that we mask the occurrences of $w$ for obtaining the mention vectors. This has two important advantages. First, it allows us to specifically capture what each sentence reveals about $w$. %\steven{For instance,} if the sentence is non-informative, then the corresponding vector should ideally be non-informative as well. 
In particular, $\mathbf{w}_{\textsc{Avg}}$ reflects the properties of $w$ that can be inferred from typical sentences mentioning this word, rather than the properties that best discriminate $w$ from other words. The result is that the vectors $\mathbf{w}_{\textsc{Avg}}$ are qualitatively different from the vectors that are obtained by standard word embedding models, as we will see in the experiments in Section \ref{secEvaluation}. Second, since we replace $w$ by a single [MASK] token, we always obtain a single vector, even if $w$ corresponds to multiple word-pieces. In contrast, without masking, the predictions for the different word-pieces from the same word have to be aggregated in some way. %(e.g.\ by averaging them). %In Section \ref{secEvaluation}, we will empirically analyse the advantage of using masking to obtain mention vectors. 

\paragraph{Filtering Idiosyncratic Mention Vectors}
%\paragraph{Removing Outliers.}
% A possible strategy for improving the representation $\mathbf{w}_{\textsc{Avg}}$, based on the second observation above, consists in removing those vectors whose cosine similarity with the static word vectors is lowest.
%In particular, let us write $\mathbf{u}_{\textsc{static}}$ for the average of the static word vectors, i.e.:
%$$
%\mathbf{u}_{\textsc{static}} = \text{Avg}(\{\mathbf{w}_{\textsc{static}} \,|\, w\in V_{\textsc{BERT}}\})
%$$
%Let $f_{\rho}(X)$, for $\rho\in [0,100]$ and $X$ a set of mention vectors, be the set of vectors obtained from $X$ after removing the $\rho\%$ least similar vectors to $\mathbf{u}_{\textsc{static}}$.
%Then we define:
%$$
%\mathbf{w}_{\textsc{Avg}}^{\rho} = \text{Avg}(f_{\rho}(\mu(w)))
%$$
%\paragraph{Removing overly specific vectors.} 
Our aim is to learn vector representations that reflect the semantic properties of nouns. One possible strategy is to compute the average $\mathbf{w}_{\textsc{Avg}}$  of the mention vectors in $\mu(w)$. However, some of these mention vectors are likely to capture idiosyncratic properties. Our hypothesis is that including such mention vectors degrades the quality of the word representations. To test this hypothesis, we introduce a strategy for identifying idiosyncratic mention vectors.
%If word vectors emphasise the idiosyncratic properties of words, this may make it harder to characterise the commonalities between different words. It may thus be of interest to remove mention vectors that capture overly specific properties. 
 %
For each mention vector $\mathbf{m}\in \mu(w)$, we compute its $k$ nearest neighbours, in terms of cosine similarity, among the set of all mention vectors that were obtained for the vocabulary $W$, i.e.\ the set $\bigcup_{v\in W} \mu(v)$. If all these nearest neighbours belong to $\mu(w)$ then we assume that $\mathbf{m}$ is too idiosyncratic and should be removed. Indeed, this suggests that the corresponding sentence expresses a property that only applies to $w$. We then represent $w$ as the average $\mathbf{w}^{*}$ of all remaining mention vectors, i.e.\ all mention vectors from $\mu(w)$ that were not found to be idiosyncratic.

\section{Evaluation}\label{secEvaluation}
Our aim is to analyse (i) the impact of masking on the averaged mention vectors and (ii) the effectiveness of the proposed filtering strategy. In Section \ref{secPropertyInduction}, we focus on the task of predicting semantic properties of words, while Section \ref{secWordSimilarity} discusses word similarity.
% property induction tasks. %We include classification problems, where categories of nouns have to be learned, as well as regression problems, where human ratings of ordinal properties need to be predicted. 
%In Section \ref{secAnalysis}, we then analyse the differences in nature between the averaged mention vectors and the static word-piece vectors that are learned by BERT. 
%Our proposed word vectors are qualitatively different from standard word vectors. While they are better-suited for modelling semantic properties, they under-perform standard word vectors on word similarity benchmarks, which we discuss in Section \ref{secWordSimilarity}. %Some further negative results are discussed in the supplementary materials.
In Section \ref{secRuleInduction}, we then evaluate the mention vectors on the downstream task of ontology completion. Section \ref{secQualitative} presents some qualitative analysis.\footnote{Implementation and data are available at \url{https://github.com/lina-luck/rosv_ijcai21}}

\smallskip
\noindent\textbf{Vector Representations.} We use two standard word embedding models for comparison: Skip-gram (SG) \cite{DBLP:journals/corr/abs-1301-3781} and GloVe \cite{DBLP:conf/emnlp/PenningtonSM14}. In both cases, we used 300-dimensional embeddings that were trained on the English Wikipedia\footnote{We used the vectors from \url{http://vectors.nlpl.eu/repository/}.}. For the contexualised vectors, we rely on two pre-trained language models\footnote{We used \url{https://github.com/huggingface/transformers}.}: BERT-large-uncased \cite{DBLP:conf/naacl/DevlinCLT19} and RoBERTa-large \cite{DBLP:journals/corr/abs-1907-11692}.
We compare a number of different strategies for obtaining word vectors from these language models. First, we use the input vectors $\mathbf{t}_{\textsc{static}}$ (\emph{Input}). For words which are not in the word-piece vocabulary, following common practice \cite{DBLP:conf/acl/BommasaniDC20,DBLP:conf/emnlp/VulicPLGK20}, we average the input embeddings of their words-pieces. %For instance, \textit{sandpaper} is tokenized by Bert into two word-pieces $\textit{sand}$ and $\textit{\#\#paper}$. The static vector of \textit{sandpaper} is then $\mathbf{v}_{sandpaper} = \frac{1}{2} (\mathbf{v}_{sand} + \mathbf{v}_{\#\#paper})$.
As the corpus for extracting mention vectors, we use the May 2016 dump of the English Wikipedia. We considered sentences of length at most 64 words to compute mention vectors, as we found that longer sentences were often the result of sentence segmentation errors. In the experiments, we only consider nouns that are mentioned in at least 10 such sentences. For nouns that occur more than 500 times, we use a random sample of 500 mentions. 

We compare two versions of the averaged mention vectors (with masking): the average of all mention vectors (AVG$_{\textit{last}}$) and the average of those that remain after applying the filtering strategy (AVG$_{\textit{filt}}$). As a baseline filtering strategy\footnote{We have also performed initial experiments with a variant of this filtering strategy, in which we first cluster the mention vectors associated with a given noun $w$ and then use the mean of the largest cluster as the final representation. However, we found this strategy to perform poorly while also being prohibitively slow to compute.}, we also show results for a variant where mention vectors were filtered based on their distance from their mean (AVG$_{\textit{outl}}$). For this baseline, we  remove a fixed percentage of the mention vectors, where this percentage is tuned as a hyper-parameter.

We also include several variants in which mention vectors are obtained without masking. For words that consist of more than one word-piece, we average the contextualised word-piece vectors. We consider the counterparts of AVG$_{\textit{last}}$,  AVG$_{\textit{filt}}$ and AVG$_{\textit{outl}}$, which we will refer to as NM$_{\textit{last}}$,  NM$_{\textit{filt}}$ and NM$_{\textit{outl}}$ respectively. Note that these strategies only look at the final layer. Previous work has found that earlier layers sometimes yield better results on lexical semantics benchmarks \cite{DBLP:conf/acl/BommasaniDC20}. For this reason, we also include a method that chooses the best layer for a given task based on the tuning split, and then uses the representations at that  layer (NM$_{=L}$). Finally, \mycite{DBLP:conf/emnlp/VulicPLGK20} suggested to take the average of the first $\ell$ layers. We also show results for this approach, where the number of layers $\ell$ is again selected for each task based on tuning data (NM$_{\leq L}$).

%\zied{Finally, we introduce  NM$_{\textit{outl}}$ and AVG$_{\textit{outl}}$ as filtering baseline, where the mention vectors are filtered based on their distance from the centroid.  

% \begin{table}[t]
% \footnotesize
% \centering
% \begin{tabular}{lcc}
% \toprule   
% \textbf{Dataset} & \textbf{Nouns}& \textbf{Properties}\\[-.5ex]
% \midrule
% McRae feature norms    &  528  & 120 \\
% BabelNet domains    &  12477  & 28 \\
% WordNet supersenses    &  18200  & 25 \\
% Glasgow Norms    &  2950  & 9 \\
% %MT40k    &  13807  & 1 \\
% \bottomrule   
% \end{tabular}
% \caption{Overview of the considered datasets.}
% \label{datasetstat}
% \end{table}

\begin{table}[t]
\footnotesize
\centering
\begin{tabular}{llcc}
\toprule   
\textbf{Dataset} & \textbf{Type} & \textbf{Nouns}& \textbf{Classes}\\[-.5ex]
\midrule
%McRae feature norms    &  528  & 120 \\
X-McRae & Commonsense & 513 & 50 \\
CSLB & Commonsense & 635 & 395 \\
Morrow & Taxonomic & 888 & 13 \\
WN supersenses  & Taxonomic   &  18200  & 25 \\
BN domains & Topical &  12477  & 28 \\
%Anew &  1640  & 3 \\
%Glasgow Norms    &  2950  & 9 \\
%MT40k    &  13807  & 1 \\
\bottomrule   
\end{tabular}
\caption{Overview of the lexical classification datasets.}
\label{datasetstat}
\end{table}

%\subsection{Property Induction}
\subsection{Modelling Semantic Properties}
\label{secPropertyInduction}
%We consider the following property induction task: given a set of words $w_1,...,w_n$ which have some (unknown) property in common, find other words which are also likely to have that property. 
\noindent \textbf{Datasets.} We consider lexical classification benchmarks involving three types of semantic properties: commonsense properties (e.g.\ \emph{table} is made of wood), taxonomic properties (e.g.\ \emph{table} is a type of furniture) and topics or domains (e.g.\ \emph{football} is related to sports). %We also evaluate our vectors on ordinal properties (e.g.\ degree of concreteness).}
In particular, we used two datasets which are focused on commonsense properties (e.g.\ being dangerous, edible, made of metal). First, we used the extension of the McRae feature norms dataset \cite{mcrae2005semantic} that was introduced in \cite{forbes2019neural} (X-McRae\footnote{\url{https://github.com/mbforbes/physical-commonsense}}). In contrast to the original McRae feature norms, this dataset contains genuine positive and negative examples for all properties.
%First, we used the Abstract feature norms\footnote{\url{https://github.com/mbforbes/physical-commonsense}} \cite{forbes2019neural}, which is a checked version of McRae feature norms \cite{mcrae2005semantic} that focus on commonsense properties (e.g.\ being dangerous, edible, made of metal). 
We considered all properties for which at least 10 positive examples are available in the dataset, resulting in a total of 50 classes.
%\zied{which result in 50 classes.}
Second, we considered CSLB Concept Property Norms\footnote{\url{https://cslb.psychol.cam.ac.uk/propnorms}}, which is similar in spirit to the McRae feature norms dataset. For this dataset, we again limited our analysis to properties with at least 10 positive examples.
We furthermore consider two datasets that are focused on taxonomic properties. First, we use the dataset that was introduced by \mycite{morrow2005representation}, which lists instances of 13 everyday categories (e.g.\ animals, fruits, furniture, instruments). Second, we used the WordNet supersenses\footnote{\url{https://wordnet.princeton.edu/download}}, which organises nouns into broad   categories, such as person, animal and plant \cite{ciaramita-johnson-2003-supersense}. %Note that words with multiple senses can belong to different categories, e.g.\ \emph{chair} belongs to the supersenses \emph{person}, \emph{artifact} and \emph{act}. 
As a final dataset, we used the BabelNet domains\footnote{\url{http://lcl.uniroma1.it/babeldomains/}} \cite{camacho2017babeldomains}, which are domain labels of lexical entities, such as \emph{music}, \emph{language}, and \emph{medicine}.
%For the experiments with ordinal properties, we used the Glasgow norms dataset \cite{scott2019glasgow},
%\footnote{\url{https://www.ncbi.nlm.nih.gov/pmc/articles/PMC6538586/}} 
%which contains ratings for 9 different features: arousal (AROU), valence (VAL), dominance (DOM), concreteness (CNC), imageability (IMAG), familiarity (FAM), age of acquisition (AOA), semantic size (SIZE) and gender association (GEND).  \zied{Second, we have used Affective Norms for English Words (ANEW) dataset \footnote{\url{https://tomlee.wtf/2010/06/16/anew/}}, which focuses on three types of emotional reactions: valence, arousal and dominance.}
%and Clark2004\footnote{\url{http://psychonomic.org/archive/}}. 
Table \ref{datasetstat} provides some statistics about the considered datasets. 
%\todo{introduce word similarity experiments}

\begin{table}[t]
\centering
\footnotesize
\addtolength{\tabcolsep}{-4.5pt} 
\begin{tabular}{
%l@{\hspace{10pt}}
%c@{\hspace{5pt}}c@{\hspace{10pt}}
%c@{\hspace{5pt}}c@{\hspace{10pt}}
%c@{\hspace{5pt}}c@{\hspace{10pt}} 
@{}l
cc
cc
cc
cc
cc
}
\toprule  
& \multicolumn{2}{c}{\textbf{X-McRae}}
& \multicolumn{2}{c}{\textbf{CSLB}}
& \multicolumn{2}{c}{\textbf{Morrow}}
& \multicolumn{2}{c}{\textbf{WordNet}}
& \multicolumn{2}{c}{\textbf{BabelNet}}\\
\cmidrule(lr){2-3}
\cmidrule(lr){4-5}
\cmidrule(lr){6-7}
\cmidrule(lr){8-9}
\cmidrule(lr){10-11}
& MAP & F1        
& MAP & F1 
& MAP & F1 
& MAP & F1 
& MAP & F1\\
 \midrule
GloVe   & 47.7 & 40.1& 43.8 & 31.8 & 54.4 & 50.6 &  42.4 & 39.4  & 33.9 & 31.8\\  
SG  & 57.7 & 49.3 & 53.3 & 39.9& 76.6 & 61.2 & 53.2 & 53.2  & 45.5 & 39.6 \\      
Input & 69.1 & 59.2 & 56.5 & 41.4 & 54.7 & 45.9 & 33.3 & 35.9 & 29.3 & 31.8 \\ 
NM$_{\textit{last}}$ & 70.8 & 61.3 & 60.8 & 45.9 & 73.0 & 56.2 & 45.6 & 43.5 & 38.7 & 38.8\\
NM$_{\textit{outl}}$ & 68.5 & 58.6& 62.1 & 47.8 & 74.3 & 68.2 & 41.9 & 43.2 & 39.8 & 39.2\\
NM$_{\textit{filt}}$ & 45.0 & 42.2 & 49.0 & 36.0 & 62.7 & 48.8 & 31.9 & 33.6 & 31.2 & 32.8\\
NM$_{=L}$ & 70.4 & 60.4 & 62.7 & 49.9 & 78.5 & 58.5 & 46.7 & 44.6 & 43.4 & 39.3\\
NM$_{\leq L}$ & 70.6 & 60.9 & 63.3 & 48.8 & 76.4 & 62.9 & 46.6 & 44.8 & 42.6 & 40.1\\
AVG$_{\textit{last}}$& 72.5 & 62.8 & 59.3 & 46.5 & 77.5 & 67.3 & 66.5 & 60.9 & 57.4 & 52.1  \\ 
AVG$_{\textit{outl}}$& 68.1 & 61.1  & 61.3 & 49.9 & 77.2 & 66.5 & 50.9 & 50.6 & 42.3 & 41.8\\ 
AVG$_{\textit{filt}}$  & \textbf{73.0} & \textbf{64.1} & \textbf{64.4} & \textbf{50.9} & \textbf{81.7} & \textbf{70.1} &  \textbf{67.8} & \textbf{61.2}  & \textbf{57.9} & \textbf{52.5} \\ 
\bottomrule  
\end{tabular}
\addtolength{\tabcolsep}{4.5pt} 
\caption{Results ($\%$) for BERT-large-uncased on the lexical classification tasks, in terms of MAP and F1 scores ($\%$).}
\label{tabClassificationBERT}
\end{table}

\begin{table}[t]
\centering
\footnotesize
\addtolength{\tabcolsep}{-4.5pt} 
\begin{tabular}{
%l@{\hspace{10pt}}
%c@{\hspace{5pt}}c@{\hspace{10pt}}
%c@{\hspace{5pt}}c@{\hspace{10pt}}
%c@{\hspace{5pt}}c@{\hspace{10pt}} 
@{}l
cc
cc
cc
cc
cc
}
\toprule  
& \multicolumn{2}{c}{\textbf{X-McRae}}
& \multicolumn{2}{c}{\textbf{CSLB}}
& \multicolumn{2}{c}{\textbf{Morrow}}
& \multicolumn{2}{c}{\textbf{WordNet}}
& \multicolumn{2}{c}{\textbf{BabelNet}}\\
\cmidrule(lr){2-3}
\cmidrule(lr){4-5}
\cmidrule(lr){6-7}
\cmidrule(lr){8-9}
\cmidrule(lr){10-11}
& MAP & F1        
& MAP & F1 
& MAP & F1 
& MAP & F1 
& MAP & F1\\
 \midrule
GloVe   & 47.7 & 40.1& 43.8 & 31.8 & 54.4 & 50.6 &  42.4 & 39.4  & 33.9 & 31.8\\  
SG  & 57.7 & 49.3 & 53.3 & 39.9& 76.6 & 61.2 & 53.2 & 53.2  & 45.5 & 39.6 \\   
Input  & 39.9 & 34.0 & 36.6 & 25.7 & 31.5 & 32.5 & 45.6 & 43.8 & 33.0 & 30.8 \\
NM$_{\textit{last}}$  & 65.2 & 54.4 & 57.2 & 42.9 & 75.2 & 62.5  & 53.0 & 50.9 & 41.2 & 40.5 \\
NM$_{\textit{outl}}$ & 66.1 &53.7 & 58.1 & 44.4 & 78.5 & 68.9 & 53.8 & 51.6 & 42.4 & 41.8\\
NM$_{\textit{filt}}$ & 37.8 & 33.0 & 40.7 & 29.3 & 57.5 & 43.6 & 39.4 & 41.1 & 35.6 & 36.3\\
NM$_{=L}$ & 65.2 & 55.9 & 60.5 & 45.6 & 79.0 & 67.5 & 53.0 & 50.9 & 43.4 & 40.3\\
NM$_{\leq L}$ & 63.9 & 53.5 & 58.9 & 44.2 & 77.9 & 67.5 & 51.3 & 50.2 & 43.1 & 39.3\\
AVG$_{\textit{last}}$ & 72.1 & 63.9 & 54.6 & 44.8 & 72.4 & 57.8  & 62.3 & 56.5 & 53.2 & 49.3  \\
AVG$_{\textit{outl}}$& 67.8 & 60.0 & 59.1 & 47.8 & 77.7 & 66.5 & 50.9 & 50.6 & 40.9 & 41.2\\ 
AVG$_{\textit{filt}}$ & \textbf{74.3} & \textbf{64.8} & \textbf{62.2} & \textbf{51.4}& \textbf{81.8} & \textbf{73.5}  &  \textbf{63.9} & \textbf{58.5} & \textbf{54.6} & \textbf{50.9} \\
\bottomrule  
\end{tabular}
\addtolength{\tabcolsep}{4.5pt} 
\caption{Results ($\%$) for RoBERTa-large on the lexical classification tasks, in terms of MAP and F1 scores ($\%$).}
\label{tabClassificationRoBERTa}
\end{table}

\begin{table}[t]
\centering
\footnotesize
\begin{tabular}{
@{}lccc}
\toprule  
&\textbf{SemEval} & \textbf{SimLex} & \textbf{WordSim}   \\ 
\midrule
GloVe & 70.5 & 43.7 & 78.2\\
SG     & 71.1       & 40.9        & 79.3    \\ 
%GloVe     & 70.8       & 43.7        & 78.2    \\ 
Input           & 63.3       & 50.9      & 72.9        \\ 
NM$_{\textit{last}}$           & 66.5   & 58.7       & 70.9      \\ 
%NM$_{\textit{last}}$-RoBERTa          & 23.2  & 0.5     & 34.6      \\ 
NM$_{\leq L}$ ($\ell=24$)          & \textbf{78.4}   & \textbf{57.8}     & 82.6      \\ 
%NM$_{\leq L}$-RoBERTa ($\ell=24$)         & 51.3  & 46.1     & 62.2      \\ 
NM$_{\textit{filt}}$ & 51.6       &43.4  & 50.0      \\
AVG$_{\textit{last}}$    & 55.4      & 41.0    & 60.9    \\ 
AVG$_{\textit{filt}}$ ($k=5$)  & 64.0   & 42.6      & 67.6     \\
\midrule
NM$_{\leq L}$ ($\ell=24$) + SG & 76.5 & 51 & \textbf{83.5}\\
AVG$_{\textit{last}}$ + SG & 70.5 & 43.1 & 75.7\\
AVG$_{\textit{filt}}$ ($k=5$) + SG & 75.2 & 43.2 & 78.9\\
\bottomrule  
\end{tabular}
\caption{Word similarity results for BERT (Spearman correlation).}
\label{tab:similarity}
\end{table}

\begin{figure}[t]
\centering
\includegraphics[width=165pt]{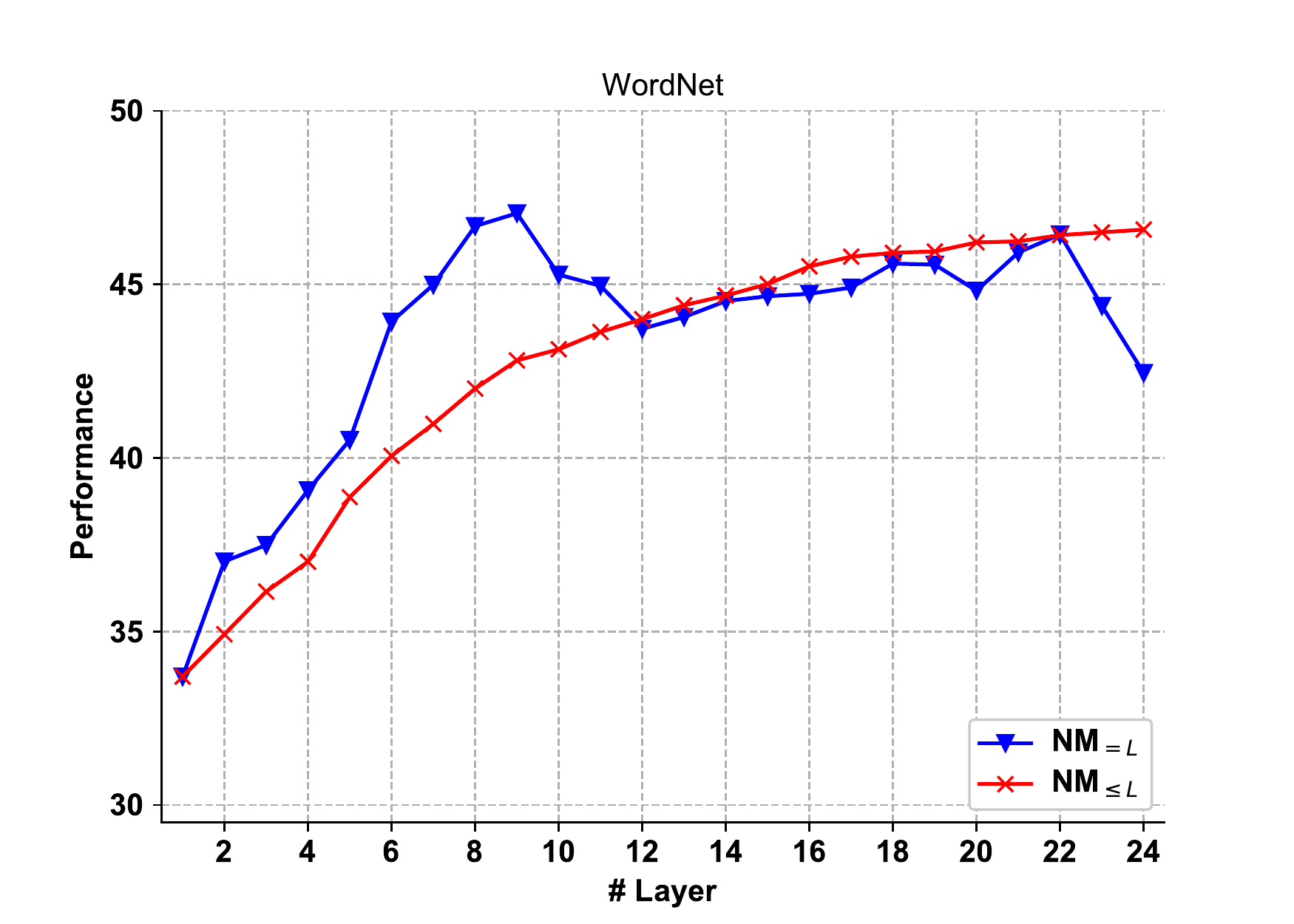}
\caption{Results of NM$_{=L}$ and NM$_{\leq L}$ per layer for WordNet supersenses with BERT.}
\label{figLayerPlot44}
\end{figure}

\begin{table}[t]
\footnotesize
\begin{tabular}{@{}l@{\hspace{3pt}}p{200pt}@{}}
\toprule
\multicolumn{2}{c}{\textbf{Top quartile gold pairs ranked in bottom quartile}} \\         
\midrule
\textbf{SimLex} &
(right, justice), \textcolor{blue}{(adult, guardian)}, \textcolor{blue}{(bird, turkey)}, (bubble, suds), (crowd, bunch), \textcolor{blue}{(flower, violet)}, (alcohol, cocktail), (evening, dusk), (wisdom, intelligence), \textcolor{blue}{(marijuana, herb)}, (intelligence, logic), (sofa, chair), (communication, language), \textcolor{blue}{(violin, instrument)}, \textcolor{blue}{(politician, president)}, \textcolor{blue}{(rabbi, minister)}\\
\midrule
\textbf{SemEval} &
(cell, lock-up), (can, bottle), \textcolor{blue}{(coca-cola, coke)}, \textcolor{green!70!black}{(obama, clinton)}, (renaissance, renascence), \textcolor{blue}{(amazon, forest)}, (english, american), \textcolor{green!70!black}{(mercury, jupiter)}, \textcolor{green!70!black}{(mercedes, opel)}, \textcolor{green!70!black}{(plato, aristotle)}, \textcolor{blue}{(orion, constellation)}, \textcolor{green!70!black}{(playstation, wii)}, \textcolor{green!70!black}{(nike, adidas)}, \textcolor{green!70!black}{(backgammon, go)}, \textcolor{green!70!black}{(kfc, mcdonald's)}, \textcolor{green!70!black}{(guardian, times)}, \textcolor{green!70!black}{(paint, photoshop)}, \textcolor{green!70!black}{(jpeg, pdf)}, (bible, gospel), \textcolor{green!70!black}{(gauss, scientist)}, (chart, graph)\\
\midrule
\textbf{WordSim} &
\textcolor{green!70!black}{(mexico, brazil)}, (dollar, buck), \textcolor{green!70!black}{(harvard, yale)}, (cell, phone)\\[0.3em]
\toprule
\multicolumn{2}{c}{\textbf{Bottom quartile gold pairs ranked in top quartile}} \\         
\midrule
\textbf{SimLex} &
(meat, bread), \textcolor{red}{(winter, summer)}, (dog, cat), \textcolor{red}{(floor, ceiling)}, \textcolor{red}{(south, north)}, \textcolor{red}{(absence, presence)}, (chocolate, pie), (bread, cheese), (river, valley), \textcolor{red}{(sunset, sunrise)}, (dog, horse), (cat, rabbit), (lawyer, banker), (wife, husband), \textcolor{red}{(bottom, top)}, (mouse, cat), (sun, sky)\\
\midrule
\textbf{SemEval} &
(gravity, meteor), (wood, blanket)\\
\midrule
\textbf{WordSim} &
(lad, wizard)\\
\bottomrule
\end{tabular}
\caption{Analysis of the similarity results for AVG$_{\textit{filt}}$.\label{tabQualitativeSim}}
\end{table}

\smallskip
\noindent \textbf{Experimental Setup.}
For all datasets, we train a binary linear SVM classifier for each of the associated classes.
%To model these properties, we train an SVM classifier.
In the case of X-McRae, we used the standard training and test splits that were provided as part of this dataset. As no validation set was provided, we reserved 20\% of the training split for hyper-parameter tuning.  
%\todo{Does the dataset contain a standard validation split as well?  --> no only training and test splits 
%If not, how did you do tuning for this experiment? --> we considered 20\% from the training as the other datasets}. 
For the remaining four datasets, we randomly split the positive examples, for each class, into 60\% for training, 20\% for tuning and 20\% for testing. Since these datasets do not provide explicit negative examples, for the test set we randomly select words from the other classes as negative examples (excluding words that also belong to the target category). The number of negative test examples was chosen as 5 times the number of positive examples. 
For the training and tuning sets, we randomly select nouns from the BERT vocabulary as negative examples. %This means that we treat the considered task as a property induction problem, where we are only given positive examples (i.e.\ the task is made harder by the fact that negative test examples come from a different distribution). % (to ensure that we can use the same training examples for Input). 
The number of negative examples for training was set as twice the number of positive examples. %For testing, as negative examples, we select words from the considered dataset which are not asserted to have the property. 
We report results of the SVM classifiers in terms of F1 score and Mean Average Precision (MAP). The latter treats the problem as a ranking task rather than a classification task, which is motivated by the fact that finding the precise classification boundary is difficult without true negative examples.  %, which is captured using MAP. %For the ordinal properties, we again split the nouns into 60\% for training, 20\% for tuning and 20\% for testing. In this case, we train a Support Vector Regression (SVR) model, which we evaluate using  Spearman $\rho$ and Kendall $\tau$.
Details about hyper-parameter tuning can be found in the supplementary materials.

%For the threshold of the classification problem, we apply greedy search based on F1 score. We need to say which parameter values were considered during tuning.} . 
%\zied{We also experiment with two different kernels: linear and RBF. In fact, using different kernels can reveal how properties are captured}. For instance, if good results are obtained with a linear kernel, this suggests that the properties can be modelled as vectors. %Similarly, good results with a quadratic kernel suggest that the words that have the given property (to a particular extent) are clustered together in the space. 
%\zied{In this paper, we only show results for linear kernel as we found out that it performs almost identically to the RBF kernel. The result for RBF kernel can be found in the supplementary materials.}

\smallskip
\noindent\textbf{Results.} The results are shown in Table \ref{tabClassificationBERT} for BERT and in Table \ref{tabClassificationRoBERTa} for RoBERTa. %In the supplementary materials, we also show results for a RBF kernel, which are highly similar to those of the linear kernel.} 
In all cases, we find that the best results are obtained for the averaged mention vectors with our proposed filtering strategy (AVG$_{\textit{filt}}$). 
The baseline filtering strategy (AVG$_{\textit{outl}}$) is clearly not competitive, leading to worse results than AVG$_{\textit{last}}$ in most cases. Another clear observation is that our proposed filtering strategy is clearly unsuitable for the NM vectors. This is because without masking, the mention vectors are clustered per word type \cite{mickus2019you}, hence the filtering strategy simply removes the majority of the mention vectors in this case; e.g.\ for X-McRae, without masking 83.1\% of the BERT mention vectors are filtered, compared to 39.7\% with masking.
Overall, strategies with masking outperform those without masking, often substantially. Among the strategies without masking, NM$_{=L}$ and NM$_{\leq L}$ perform best, confirming the finding from earlier work that the last layer is not always optimal \cite{DBLP:conf/acl/BommasaniDC20}, although in contrast to \mycite{DBLP:conf/emnlp/VulicPLGK20} we do not find that NM$_{\leq L}$ clearly outperforms NM$_{=L}$. For NM$_{\leq L}$ we find that $\ell=24$ is chosen for most of the cases, whereas for NM$_{= L}$ layers 8-10 are often best. Figure \ref{figLayerPlot44} shows the performance per layer for WordNet supersenses. Layer-wise results for all datasets are provided in the supplementary materials.

\subsection{Word Similarity}\label{secWordSimilarity}
We now consider word similarity benchmarks, where the task consists in ranking word pairs based on their degree of similarity. This ranking is then compared with a gold standard obtained from human judgements. We consider three standard word similarity datasets for nouns: the similarity portion of WordSim \cite{Agirreetal:09}, the noun subset of SimLex \cite{hill-etal-2015-simlex} and SemEval-17 \cite{camacho-collados-etal-2017-semeval}. Many word similarity datasets, such as MEN \cite{bruni2014multimodal}, RG-65 \cite{RG65:1965} or the full WordSim-353 \cite{Levetal:2002}, measure relatedness (or degree of association). In contrast, the  datasets that we consider here all mainly focus on similarity. In particular, SimLex does not consider relatedness at all in the similarity scale, while SemEval and WordSim only consider relatedness in the lower grades of their similarity scales.

Table \ref{tab:similarity} shows the results of the word similarity experiments (for BERT). For SimLex, which is least affected by relatedness, all BERT based representations outperform SG, but the masking based strategies underperform GloVe. For the two other datasets, the masking based strategies underperform the SG and GloVe baselines. In all cases, the filtering strategy AVG$_{\textit{filt}}$ improves the results of AVG$_{\textit{last}}$, but the best results are consistently found without masking. 
To analyse the complementarity of the representations, we also experiment with strategies where two different representations are concatenated (after normalising the vectors). For AVG$_{\textit{filt}}$ this considerably improves the results, which shows that SG and AVG$_{\textit{filt}}$ capture complementary aspects of similarity.
To better understand the reasons for the under-performance of AVG$_{\textit{filt}}$, Table \ref{tabQualitativeSim} shows all pairs that are within the top quartile of most similar pairs according to the gold ratings while being in the bottom quartile according to the AVG$_{\textit{filt}}$ vector similarity, and vice versa. As can be seen, the pairs with high gold ratings but low vector similarity include several hypernym pairs (shown in blue) and named entities (shown in green). Conversely, the pairs with low gold ratings but high vector similarity include mostly co-hyponyms and antonyms\footnote{It should be noted that antonyms are purposely given low gold scores in SimLex as per their annotation guidelines.} (shown in red). 

%****************************

\begin{table}[t]
\centering
\footnotesize
\begin{tabular}{
l
ccc
ccc
ccc
ccc
ccc
}
\toprule  
& \multicolumn{1}{c}{\textbf{Wine}}
& \multicolumn{1}{c}{\textbf{Econ}}
& \multicolumn{1}{c}{\textbf{Olym}}
& \multicolumn{1}{c}{\textbf{Tran}}
& \multicolumn{1}{c}{\textbf{SUMO}}
\\
%\cmidrule(lr){2-4}
%\cmidrule(lr){5-7}
%\cmidrule(lr){8-10}
%cmidrule(lr){11-13}
%\cmidrule(lr){14-16}
%& 
%& Pr & Rec & F1 
%& Pr & Rec & F1
%& Pr & Rec & F1
%& Pr & Rec & F1\\
 \midrule
 SG & 13.8 &  13.5 & 8.3 & 7.2 & 33.4\\  
 Input & 22.2 &  14.2 & 12.1 & 9.4 & 36.5\\
 NM$_{\textit{last}}$ & 18.5 &  12.5 & 16.2 & 12.3 &  38.9\\
NM$_{\textit{filt}}$ & 20.3 &  15.6 & 13.3 &  11.2 & 35.4\\
AVG$_{\textit{last}}$ & 23.0 &20.0  & 16.9 &  11.5 &  41.4 \\
AVG$_{\textit{filt}}$ & \textbf{24.5} &  \textbf{24.3} & \textbf{22.9}&  \textbf{13.0} & \textbf{46.4}\\
\bottomrule 
\end{tabular}

%\begin{tabular}{
%@{}l
%ccc
%ccc
%ccc
%ccc
%ccc
%}
%\toprule  
%& \multicolumn{3}{c}{\textbf{Wine}}
%& \multicolumn{3}{c}{\textbf{Economy}}
%& \multicolumn{3}{c}{\textbf{Olympics}}
%& \multicolumn{3}{c}{\textbf{Transport}}
%& \multicolumn{3}{c}{\textbf{SUMO}}
%\\
%\cmidrule(lr){2-4}
%\cmidrule(lr){5-7}
%\cmidrule(lr){8-10}
%\cmidrule(lr){11-13}
%\cmidrule(lr){14-16}
%& Pr & Rec & F1        
%& Pr & Rec & F1 
%& Pr & Rec & F1
%& Pr & Rec & F1
%& Pr & Rec & F1\\
% \midrule
% SG & 19.6 & 14.3 & 13.8 & 36.1 & 11.8 & 13.5 & 16.7 & 5.6 & 8.3 & 16.9 & 8.1 & 7.2 & 39.1 & 29.3 &33.4\\  
% Input & 29.3  & 28.7 & 22.2 & 33.1 & 12.9 & 14.2 & 13.3 & 11.1 & 12.1 & 34.7 & 5.9 & 9.4 & 42.7 &31.9 &36.5\\
% NM$_{\textit{last}}$ & 34.1 & 20.8 & 18.5 & 28.3 & 9.0 & 12.5 & 50.0 & 9.7 & 16.2 & 39.8 & 8.5 &  12.3 & 44.5 & 34.6 & 38.9\\
% NM$_{\textit{filt}}$ & 38.2 & 32.6 & 20.3 & 50.0 & 14.3 & 15.6 & 8.8 & 27.8 & 13.3 & 33.7 & 7.1 & 11.2 & 41.9 & 30.7 & 35.4\\
% AVG$_{\textit{last}}$ & 30.7  & 23.8 & 23.0 & 39.2 & 18.9 & 20.0  & 66.7 & 9.7 & 16.9 & 37.2 & 10.0 & 11.5 & 46.6& 37.3& 41.4 \\
% AVG$_{\textit{filt}}$ & \textbf{34.7} & \textbf{33.5} & \textbf{24.5} & \textbf{69.3} & \textbf{15.2} & \textbf{24.3} & \textbf{44.4} & \textbf{22.2} & \textbf{22.9}& \textbf{41.0} & \textbf{11.0} & \textbf{13.0} & \textbf{51.3} & \textbf{42.5} & \textbf{46.4}\\
%\bottomrule 
%\end{tabular}
\caption{Results ($\%$ F1) for ontology completion for BERT.}
\label{tabOntologyCompletion}
\end{table}

\begin{table}[t]
%\renewcommand{\arraystretch}{1.15}
%\resizebox{\textwidth}{!}{
\centering
\footnotesize
\addtolength{\tabcolsep}{-3pt} 
\begin{tabular}{@{}ll@{}}
\toprule
              & \multicolumn{1}{c}{\textbf{saint}}                                                                                                           \\ \cline{2-2} 
%Input          & sainthood, saintliness, st, stag, stob, strum, stamen \\ %, stavanger, stipe,  sty                                                        \\
NM$_{\textit{last}}$         & st, sainthood, saintliness, stob, strontianite, sanctuary \\%, stag, postulant, church, martyr                                          \\
AVG$_{\textit{last}}$         & st, pope, monsieur, prince, martyr, sage, antipope \\ %, monsignor, mons, archangel                                                      \\
AVG$_{\textit{filt}}$         & martyr, bishop, archangel, sage, patriarch, deacon \\%, confessor, druid, beguine, prophet                                              \\ 
\midrule
              & \multicolumn{1}{c}{\textbf{emeritus}}                                                                                                        \\ \cline{2-2} 
%Input           & laureate, episcopalianism, alumnus, professorship \\ %, dowager, toxicologist, sporotrichosis, thyroeidectomy,  ...\\ % trusteeship, aromatherapy \\
NM$_{\textit{last}}$          & adviser, incumbent, appointment, retirement, honorarium \\ %, academician, executive, advisory, ...\\ % alumnus, expert                          \\
AVG$_{\textit{last}}$          & dean, visiting, hod, excellency, chair, professor, assistant \\ %, executive, provost, docent                                            \\
AVG$_{\textit{filt}}$          & fellow, laureate, excellency, provost, principal, hod \\%, lectureship, diplomate, supremo, dean                                        \\  
\midrule
              & \multicolumn{1}{c}{\textbf{rent}}                                                                                                            \\ \cline{2-2} 
%Input           & renter, rentier, rental, lease, leasehold, landlord, tenant \\ %,  abetalipoproteinemia, mascarpone,  costochondritis                       \\
NM$_{\textit{last}}$          & rentier, rental, lease, tenant, landlord, renter, leasehold \\%, fee, fare, bailment                                                    \\
AVG$_{\textit{last}}$          & lease, rental, royalty, mortgage, scrip, wage, cash \\%, purchase, payroll, tax                                                          \\
AVG$_{\textit{filt}}$          & lease, mortgage, purchase, loan, expense, royalty, debt \\ %, dividend, allowance, refund                                                \\  
\midrule
%              & \multicolumn{1}{c}{\textbf{donor}}                                                                                                           \\ \cline{2-2} 
%Input           & benefactor, philanthropy, mineralocorticoid, recipient, philanthropist, oxidoreductase, hyperlipopoteinemia ... \\
%chapultepec                     \\
%NM$_{\textit{last}}$          & benefactor, philanthropist, recipient, charity, client, gift, source, collector, contribution, philanthropy                                                      \\
%AVG$_{\textit{last}}$          & client, appropriator, participant, constituent, buyer, investor, employer, partner, benefactor, taxpayer                                                    \\
%AVG$_{\textit{filt}}$          & grantee, client, employer, appropriator, constituent, partner, moderator, lobbyist, collector, agent                                                     \\  
%\midrule
              & \multicolumn{1}{c}{\textbf{austrian}}                                                                                                          \\ \cline{2-2} 
%Input         & austria, hungarian, habsburg, bavarian, vienna, innsbruck \\%, austronesian, bolivian, belgian, croatian                                     \\
NM$_{\textit{last}}$     & austria, vienna, archduke, wiener, innsbruck, graz \\ %, habsburg, salzburg, bavaria, tyrol                                                  \\
AVG$_{\textit{last}}$       & slovak, czech, dutch, hungarian, brazilian, russian\\ %, moroccan, romanian, polish, seychellois                                                \\
AVG$_{\textit{filt}}$  & bavarian, dutch, slovak, belgian, russian, canadian\\ %, german \\ %, sicilian, neapolitan, moroccan               \\               
\bottomrule                    
\end{tabular}
\addtolength{\tabcolsep}{3pt} 
%}
\caption{Nearest neighbours for selected target words, in terms of cosine similarity, for the vocabulary from WordNet supersenses.}
\label{tab:nns}
\end{table}

\begin{table}[t]
\centering
\footnotesize
\begin{tabular}{@{}p{35pt}p{200pt}@{}}
\toprule
\textbf{Target} & \textbf{Masked sentence}\\
\midrule
banana & Some countries produce statistics distinguishing between \underline{\phantom{aaa}} and plantain production, but four of ... \\
\midrule
sardine & Traditional fisheries for anchovies and \underline{\phantom{aaa}} also have operated in the Pacific, the Mediterranean, and ... \\
\midrule
lamb & Edison's 1877 tinfoil recording of Mary Had a Little \underline{\phantom{aaa}}, not preserved, has been called the first ...\\
\midrule
pineapple & In October 2000, the Big \underline{\phantom{aaa}}, a tourist attraction on the Sunshine Coast, was used as a backdrop for ...\\
\midrule
salamander & The southern red-backed salamander (Plethodon serratus) is a species of \underline{\phantom{aaa}} endemic to the United States.\\
\bottomrule
\end{tabular}
\caption{Examples of sentences whose corresponding mention vectors were filtered. \label{tabFilteredSentences}}
\end{table}

%**************************************
\subsection{Ontology Completion}
\label{secRuleInduction}
We now consider the downstream task of ontology completion. Given a set of ontological rules, the aim is to predict plausible missing rules \cite{li2019ontology}; e.g.\ suppose the ontology contains the following rules:
\begin{align*}
\emph{Beer}(x) &\rightarrow
\emph{AlcoholicBeverage}(x)\\
\emph{Gin}(x) &\rightarrow \emph{AlcoholicBeverage}(x)
\end{align*}
As these rules have the same structure, they define a so-called rule template of the form 
$$
\star(x) \rightarrow
\emph{AlcoholicBeverage}(x)
$$
where $\star$ is a placeholder. Since substituting the placeholder by \emph{beer} and \emph{gin} makes the rule valid, and since \emph{wine} shares most of the semantic properties that \emph{beer} and \emph{gin} have in common, intuitively it seems plausible that substituting $\star$ by \emph{wine} should also produce a valid rule.
To predict plausible rules in this way, \mycite{li2019ontology} used a graph-based representation of the rules. The nodes of this graph correspond to concepts (or predicates) while edges capture different types of interactions, derived from the rules. The predictions are made by a Graph Convolutional Network,  where skip-gram embeddings of the concept names are used as the input node embeddings.  
%The edges represent the different interactions between the concepts in the given rules. %One possibility to initialize the nodes is to consider vectors from a pre-trained word embedding. 
In this experiment, we replace the skip-gram vectors by our averaged mention vectors and evaluate the resulting predictions on four well-known domain-specific ontologies (i.e.\ Wine, Economy, Olympics and Transport) and on the open-domain ontology SUMO\footnote{\url{http://www.adampease.org/OP/}}. We used the pre-processed versions of these ontologies, and corresponding training and test splits, from \mycite{li2019ontology}\footnote{\url{https://github.com/bzdt/GCN-based-Ontology-Completion}}. As our focus is on evaluating the usefulness of the word vectors, we only generate templates for concepts whose names occur at least two times in Wikipedia. 
Furthermore, as the hyper-parameters of the GCN model are sensitive to the dimensionality of the input representations, we use Singular Value Decomposition to reduce the dimensionality of our vector representations from 1024 to 300, allowing us to keep the same hyper-parameter values as for the skip-gram vectors.
%Following \mycite{li2019ontology}, the results are evaluated in terms of precision, recall and F1 score. 
For more details about the experimental methodology, we refer to \cite{li2019ontology}. As this ontology completion model is computationally expensive, we restrict the set of baselines for this experiment, and show results for BERT only. The results are presented in Table \ref{tabOntologyCompletion}, in terms of 
%precision, recall and
F1
%. These results 
%and 
confirm that the average mention vectors considerably outperform skip-gram vectors, and that the filtering strategy leads to further improvements.

%**************************************
\subsection{Qualitative Analysis}\label{secQualitative}

%\noindent \todo{Nearest neighbours}\\

\textbf{Nearest neighbours.} In Table \ref{tab:nns}, we show the nearest neighbours of four selected words. %Our focus here is on qualitatively analysing the effect of the filtering strategy and the differences between masking (AVG$_{\textit{last}}$) and not masking (NM$_{\textit{last}}$) the target words. 
Some of the listed examples clearly illustrate how the proposed filtering step is successful in prioritizing general semantic properties. For instance, %\emph{st}, which is the nearest neighbour of \emph{saint} for AVG$_{\textit{last}}$, disappears from the top-10 nearest neighbours after filtering, whereas other concepts which share core properties (e.g., \emph{martyr} and \emph{bishop} are two concepts that evoke the properties of person, holiness and religion), are ranked higher. 
in the case of \emph{emeritus}, filtering leads to lower ranks for university-related terms (e.g.\ \emph{dean} and \emph{chair}) and higher ranks for honorary positions (e.g. \emph{fellow} and \emph{excellency}). For \emph{rent}, the filtering strategy increases the rank of concepts related to monetary transactions (e.g.\ \emph{``purchase''} and \emph{``expense''}). The effect of masking can be clearly seen in all examples, where strategies with masking (AVG$_{\textit{last}}$ and AVG$_{\textit{filt}}$) more consistently result in neighbours of the same kind. %whereas the neighbours for NM$_{\textit{last}}$ often capture more general semantic relatedness. 
For example, for \emph{austrian}, the masked variants consistently select demonyms, whereas the neighbours for NM$_{\textit{last}}$ include various terms related to Austria.
%, whereas the masked variant's nearest neighbours are persons. The same applies to \emph{austrian}, where the masked variant concentrates all demonyms, whereas the non masked cluster also has cities like \emph{graz} or \emph{salzburg}.
%Finally, the results for Input are overall far noisier. 
The example of \emph{saint} also highlights how the need to average multiple word-piece vectors can introduce further noise, as some of the selected neighbours for NM$_{\textit{last}}$ are included because they contain \emph{st} as a word-piece, despite not being semantically related. %The supplementary materials contain further analysis about the static BERT vectors, showing that they are very different in nature from the mention vectors. Among others, we find that the static BERT vectors and averaged mention vectors lie in different, approximately orthogonal sub-spaces.

\begin{figure}[t]
\centering
\includegraphics[width=120pt]{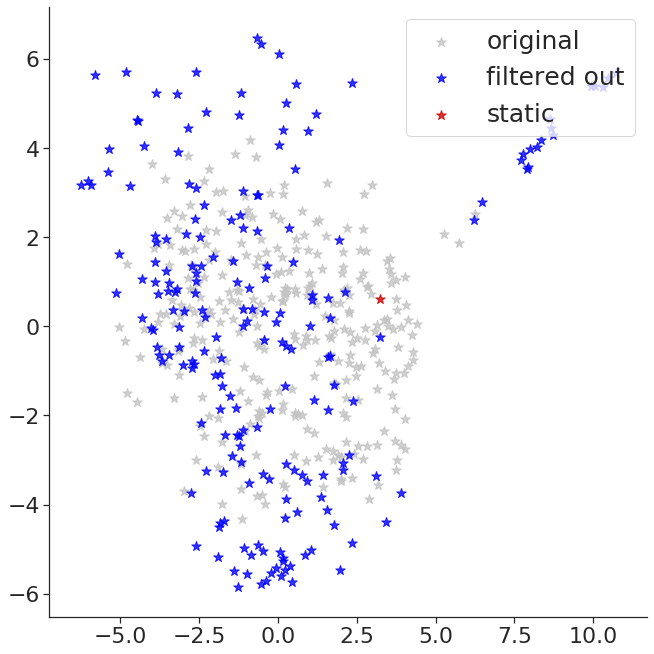}
\hfill
\includegraphics[width=120pt]{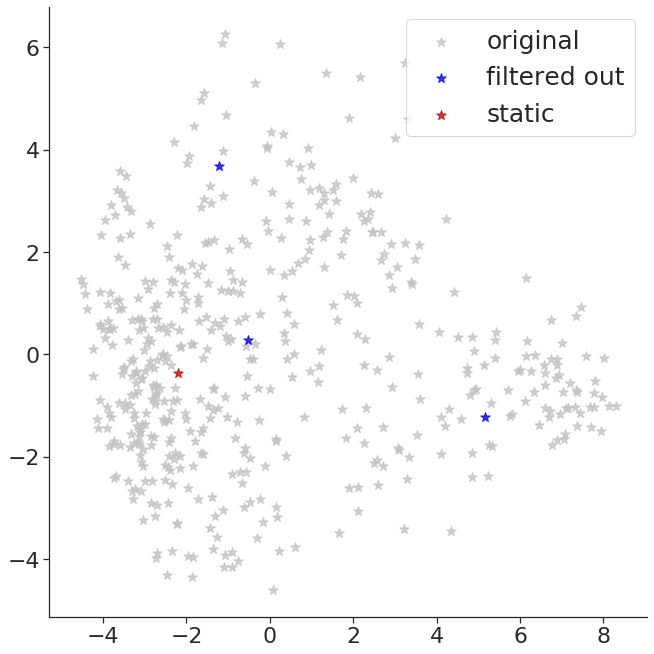}
\caption{Plots of the mention vectors for \emph{sapling} (left) and \emph{cockroach} (right) from the Morrow dataset, showing the mention vectors which are removed by the filtering strategy (blue) and the corresponding static vector (red).}
\label{figPlot}
\end{figure}

\smallskip
\noindent \textbf{Examples of filtered mentions.} Table \ref{tabFilteredSentences} provides some examples of sentences whose resulting mention vector was filtered, for words from X-McRae. %These examples illustrate several reasons why mention vectors are filtered. 
The sentence for \emph{banana} asserts a highly idiosyncratic property, namely that the words \emph{banana} and \emph{plantain} are interchangeable in some contexts. The example for \emph{sardine} is filtered because \emph{sardines} and \emph{anchovies} are often mentioned together.
The examples for \emph{lamb} and \emph{pineapple} illustrate cases where the target word is used within the name of an entity, rather than on its own. Finally, as the example for \emph{salamander} illustrates, highly idiosyncratic vectors can be obtained from sentences in which the target word is mentioned twice. To further illustrate the behaviour of the filtering strategy, Figure \ref{figPlot} depicts which mention vectors are filtered for two examples; further examples can be found in the supplementary materials. The figure shows that the strategy is adaptive, in the sense that a large number of mention vectors are filtered for some words, while only few vectors are filtered for other words. This clearly shows that our strategy is not simply removing outliers, which is in accordance with the poor performance of the AVG$_{\textit{outl}}$ baseline.

%******************************
\section{Conclusion}
%The problem of learning word vectors has already received considerable attention. However, previous work has mostly focused on the performance of such vectors in NLP tasks, where static word vectors have now largely been superseded by the use of pre-trained neural language models. In many other applications, however, word vectors remain important, because they capture prior knowledge about the commonalities between different entities (e.g.\ category labels, query terms, predicate names). It is currently less well-understood how word vectors for such applications can best be learned. 
We have analysed the potential of averaging the contextualised vectors predicted by BERT to obtain high-quality static word vectors. We found that the resulting vectors are qualitatively different depending on whether or not the target word is masked. When masking is used, the resulting vectors tend to represent words in terms of the general semantic properties they satisfy, which is useful in tasks where we have to identify words that are of the same kind, rather than merely related. We have also proposed a filtering strategy to obtain vectors that de-emphasise the idiosyncratic properties of words, leading to improved performance in the considered tasks.

%\todo{Lots of previous work on word vectors, focus has usually been on performance in NLP tasks, which is now less important. Our focus is on ``word vectors as knowledge bases''.  Different types of word vectors needed for different tasks / this paper introduced the idea of focusing on idiosyncratic vs general properties / averaged mention vectors suitable for capturing general properties when masking ; static BERT vectors and unmasked vectors suitable when needing idiosyncratic properties / proposed filtering strategy that consistently improves results in tasks that require general poperties. NM combines "word of the same kind" with overall similarity.}

% \section*{Acknowledgments}

% The acknowledgments should go immediately before the references. Do not number the acknowledgments section.
% Do not include this section when submitting your paper for review.

\section*{Acknowledgements}
This work was performed using HPC resources from GENCI-IDRIS (Grant 2021-[AD011012273]). Zied Bouraoui was funded by ANR CHAIRE IA BE4musIA.

%******************************
\appendix

\section{Experimental Details}
\paragraph{Modelling Semantic Properties.}
For the lexical classification datasets, hyper-parameters are tuned as follows.
Using the tuning splits, we select the optimal value of $k$ (i.e.\ the number of neighbours considered by our filtering strategy) from $\{3,5,10,20,50,100\}$. We also tune the parameters of the SVM  model $C \in \{0.1,\allowbreak 1,\allowbreak 10,\allowbreak 100\}$ and $\gamma\in\{0.001,\allowbreak 0.01,\allowbreak 0.1,\allowbreak 1,\allowbreak 10,\allowbreak 100,\allowbreak 1000\}$. For the NM$_{\textit{outl}}$ and AVG$_{\textit{outl}}$ baselines, the percentage of mention vectors to be removed was selected from $\{10\%,...,90\%\}$. For the NM$_{=L}$ and NM$_{\leq L}$ baselines, we also select the optimal layer $\ell \in\{1,...,24\}$ based on the tuning split.

\paragraph{Ontology Completion.}
In this section, we present more details about the experimental methodology that was followed for the ontology completion experiment. 
Given an ontology, for each concept, we extract at most $500$ sentences which mention its name, and then use these sentences to compute the averaged mention vectors. We follow the strategy from \mycite{li2019ontology} to tokenize compound concept names such as \emph{WhiteWhine}. Nonetheless, even after this tokenisation step, some concepts have names which do not tend to appear in standard text (e.g.\ \emph{PastaWithWhiteSauce} is a concept in the Wine ontology). These concepts are simply ignored in our experiments, i.e.\ we never predict such concepts as plausible instance of a given template. Accordingly, we only consider unary templates that have at least remaining concept among its instances (i.e.\ at least one of the concepts that makes the unary template valid is such that we can get mention vectors for its name). Using the remaining templates and concepts, we follow the same evaluation protocol as \mycite{li2019ontology}. In particular, we use the same hyper-parameters and use the same training and test splits.

\section{Additional Results}

\begin{figure}[p]
\centering
\includegraphics[width=200pt]{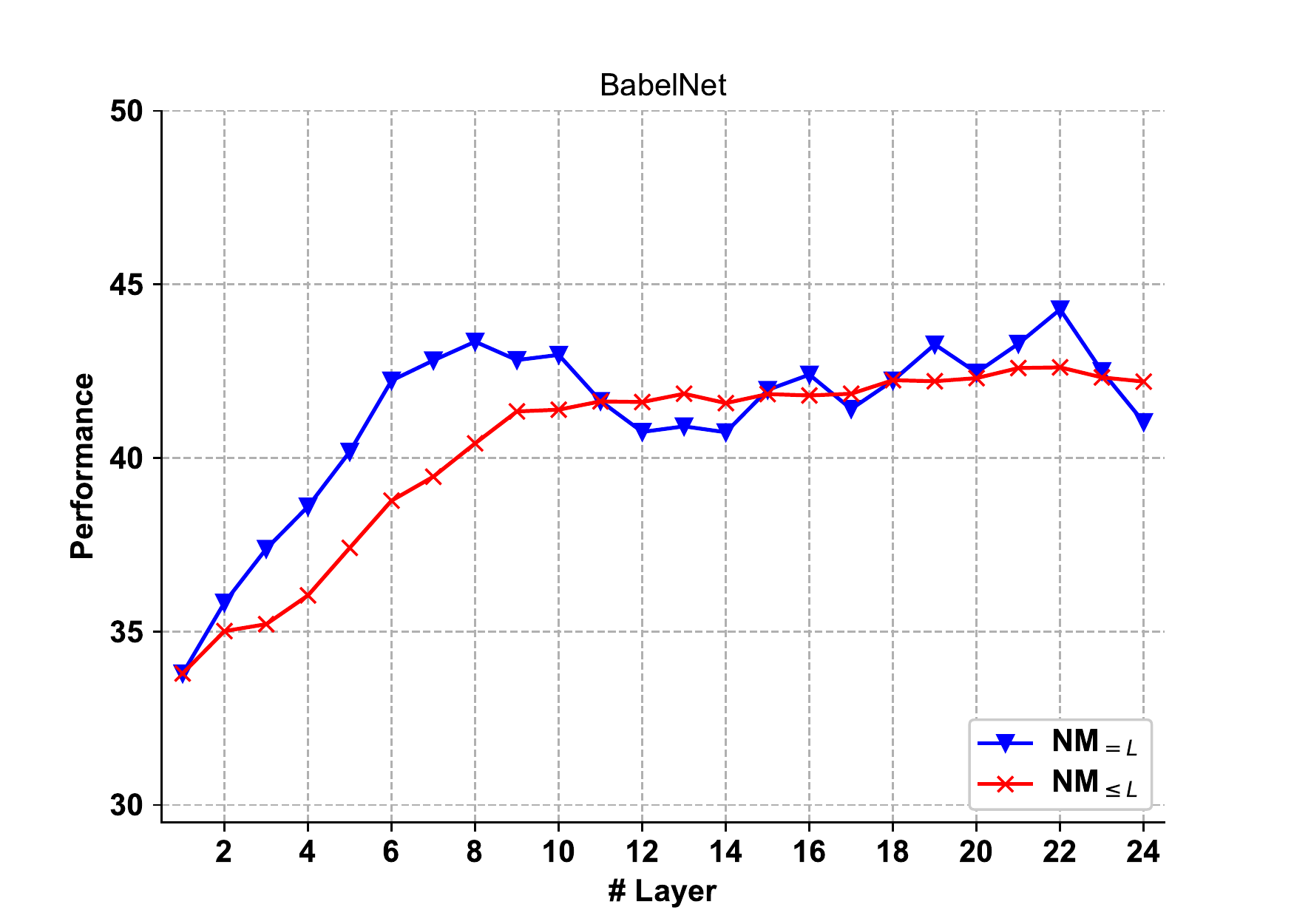}
\caption{Layer-wise performance on BabelNet domains (BERT).}
\label{figLayerPlot1}
\end{figure}

\begin{figure}[p]
\centering
\includegraphics[width=200pt]{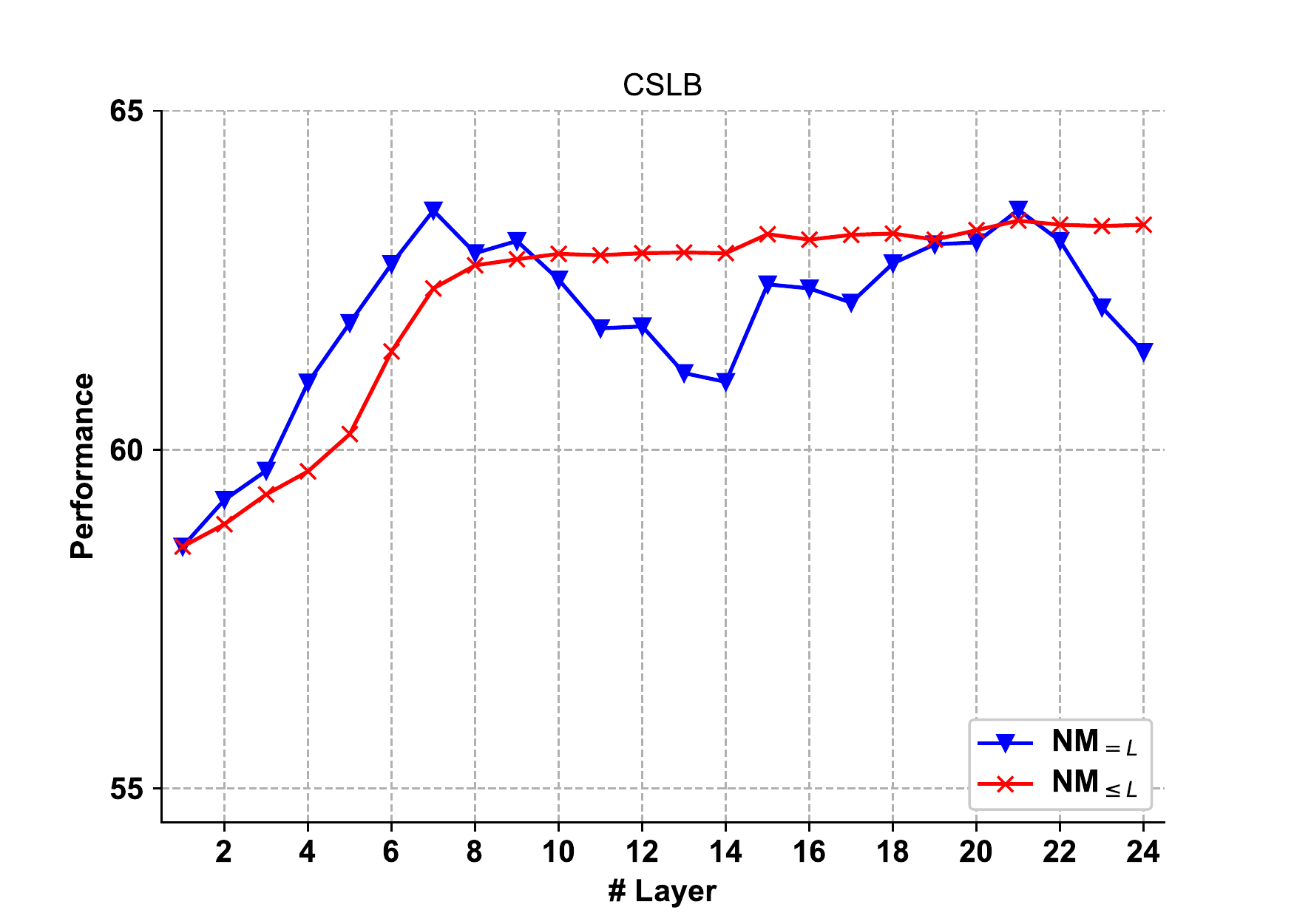}
\caption{Layer-wise performance on CSLB (BERT).}
\label{figLayerPlot2}
\end{figure}

\begin{figure}[p]
\centering
\includegraphics[width=200pt]{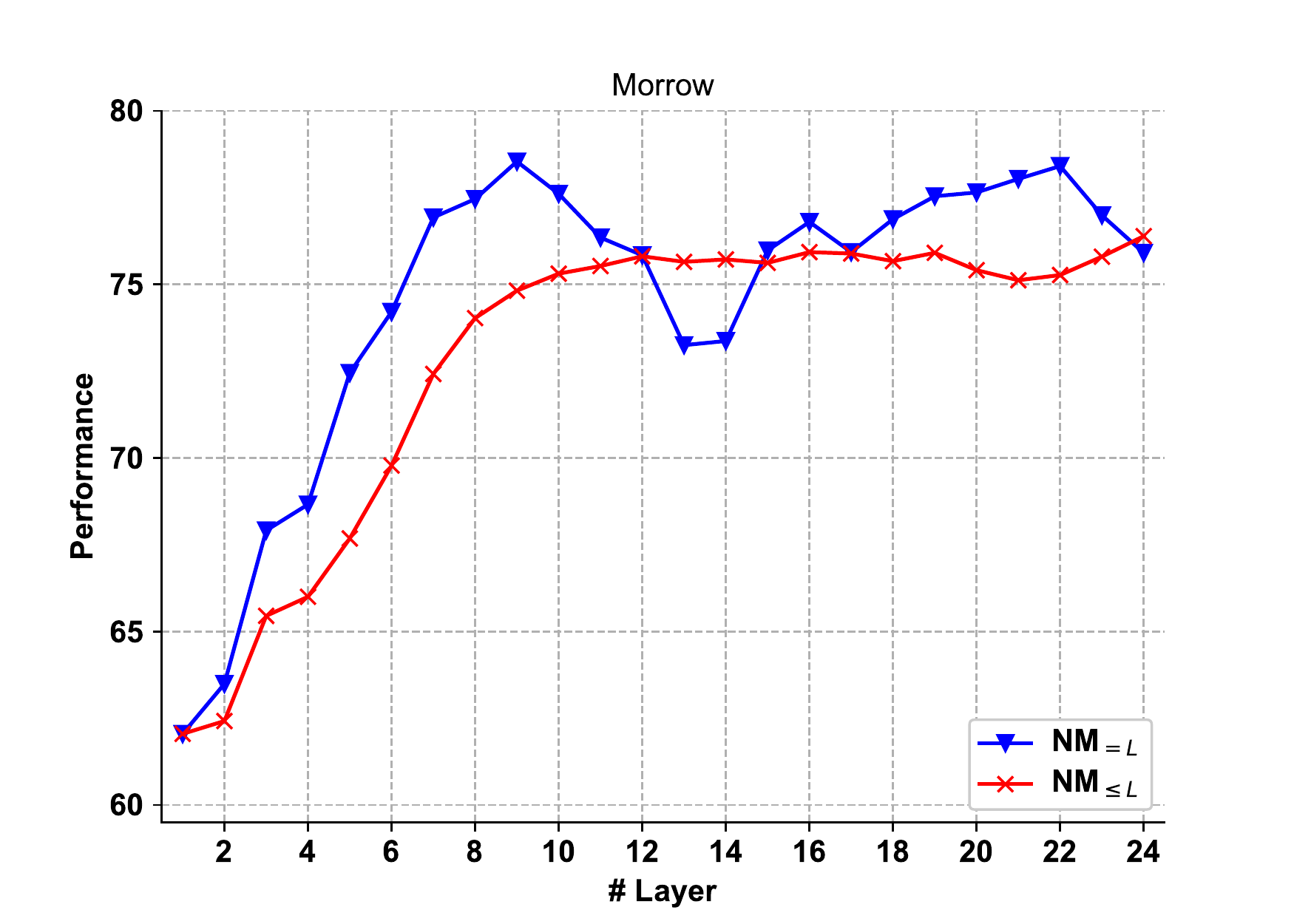}
\caption{Layer-wise performance on Morrow (BERT).}
\label{figLayerPlot3}
\end{figure}

\begin{figure}[p]
\centering
\includegraphics[width=200pt]{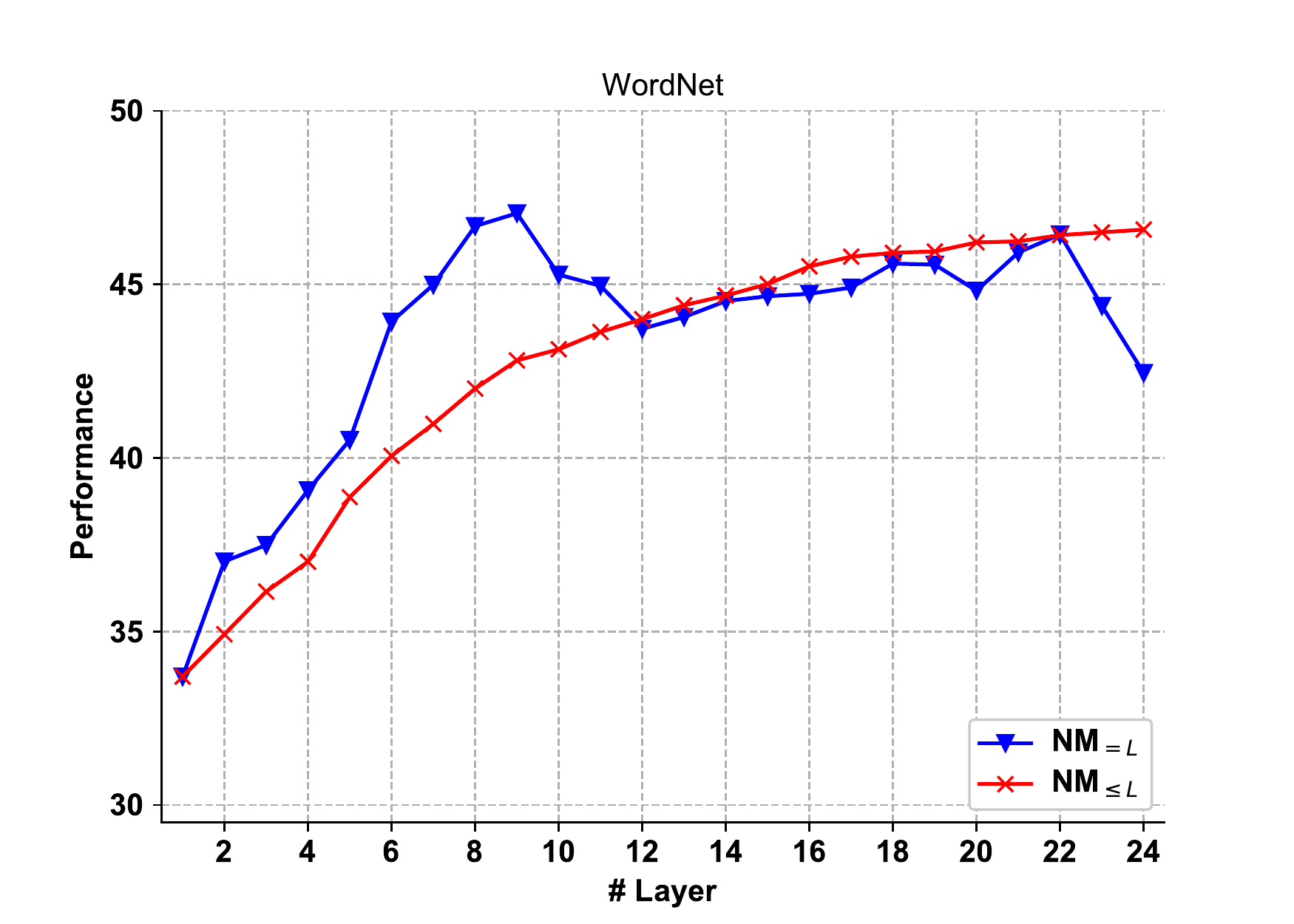}
\caption{Layer-wise performance on WordNet supersenses (BERT).}
\label{figLayerPlot4}
\end{figure}

\begin{figure}[p]
\centering
\includegraphics[width=200pt]{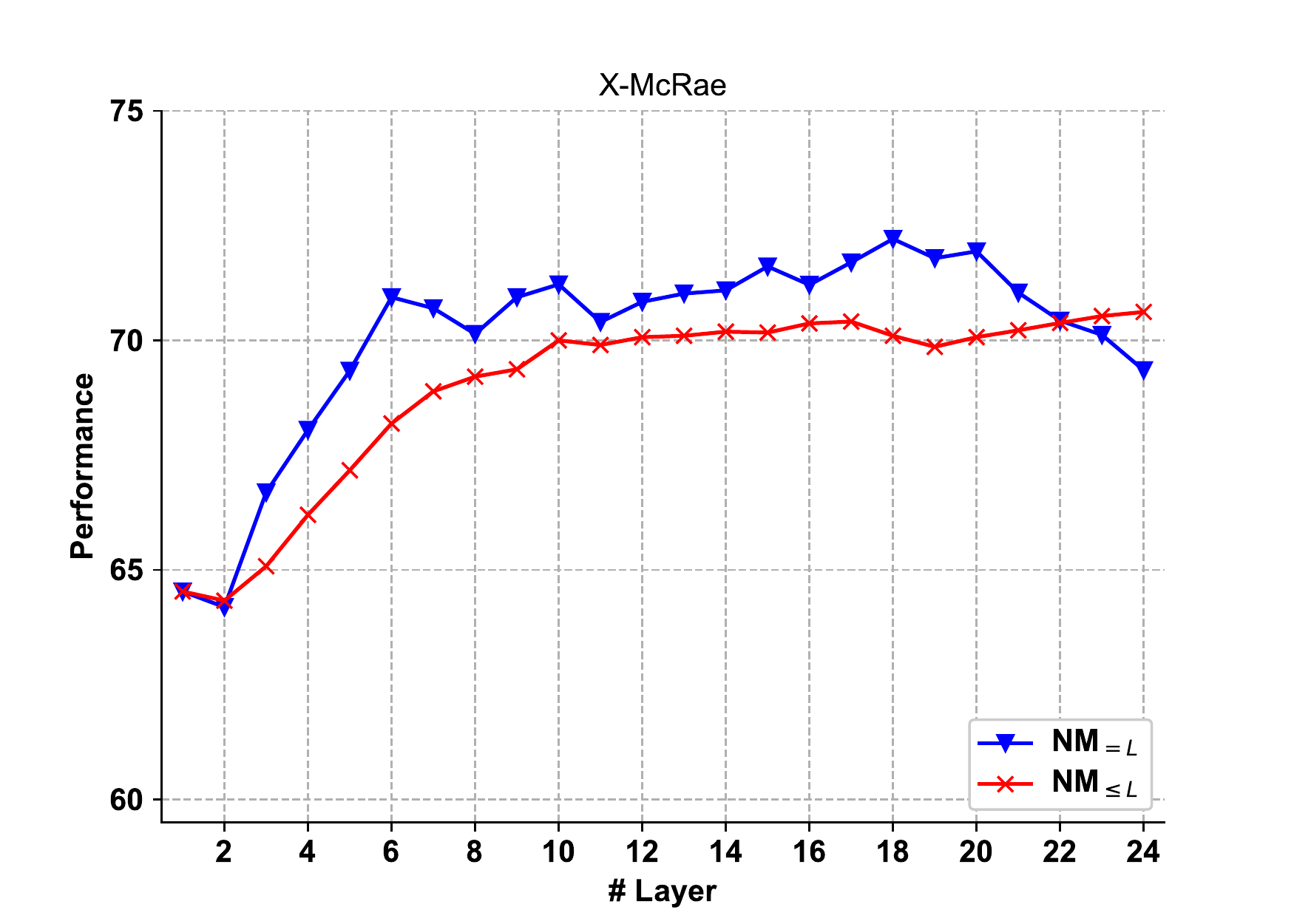}
\caption{Layer-wise performance on X-McRae (BERT).}
\label{figLayerPlot5}
\end{figure}

\begin{figure}[p]
\centering
\includegraphics[width=200pt]{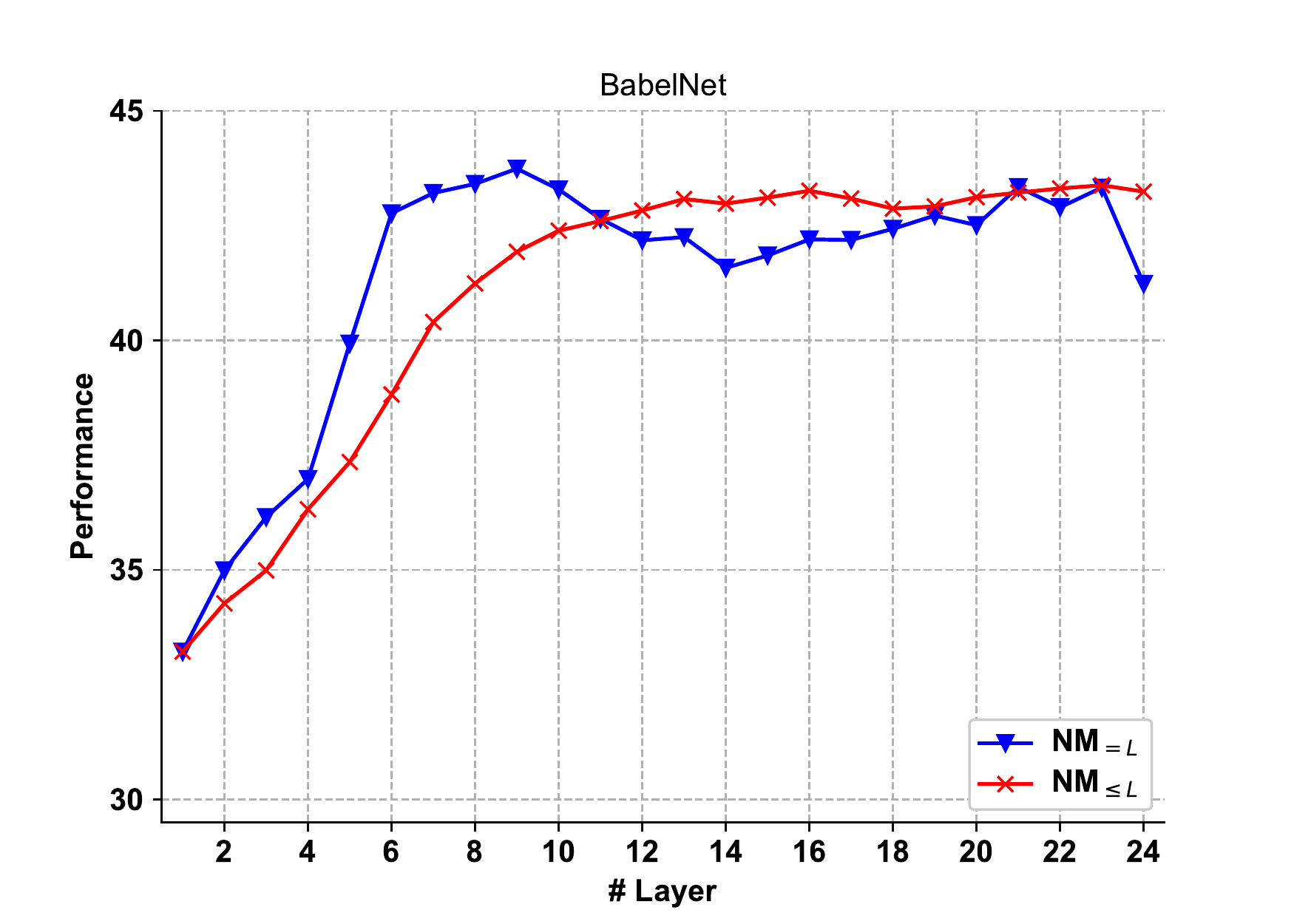}
\caption{ Layer-wise performance on BabelNet domains (RoBERTa).}
\label{figLayerPlot6}
\end{figure}

\begin{figure}[p]
\centering
\includegraphics[width=200pt]{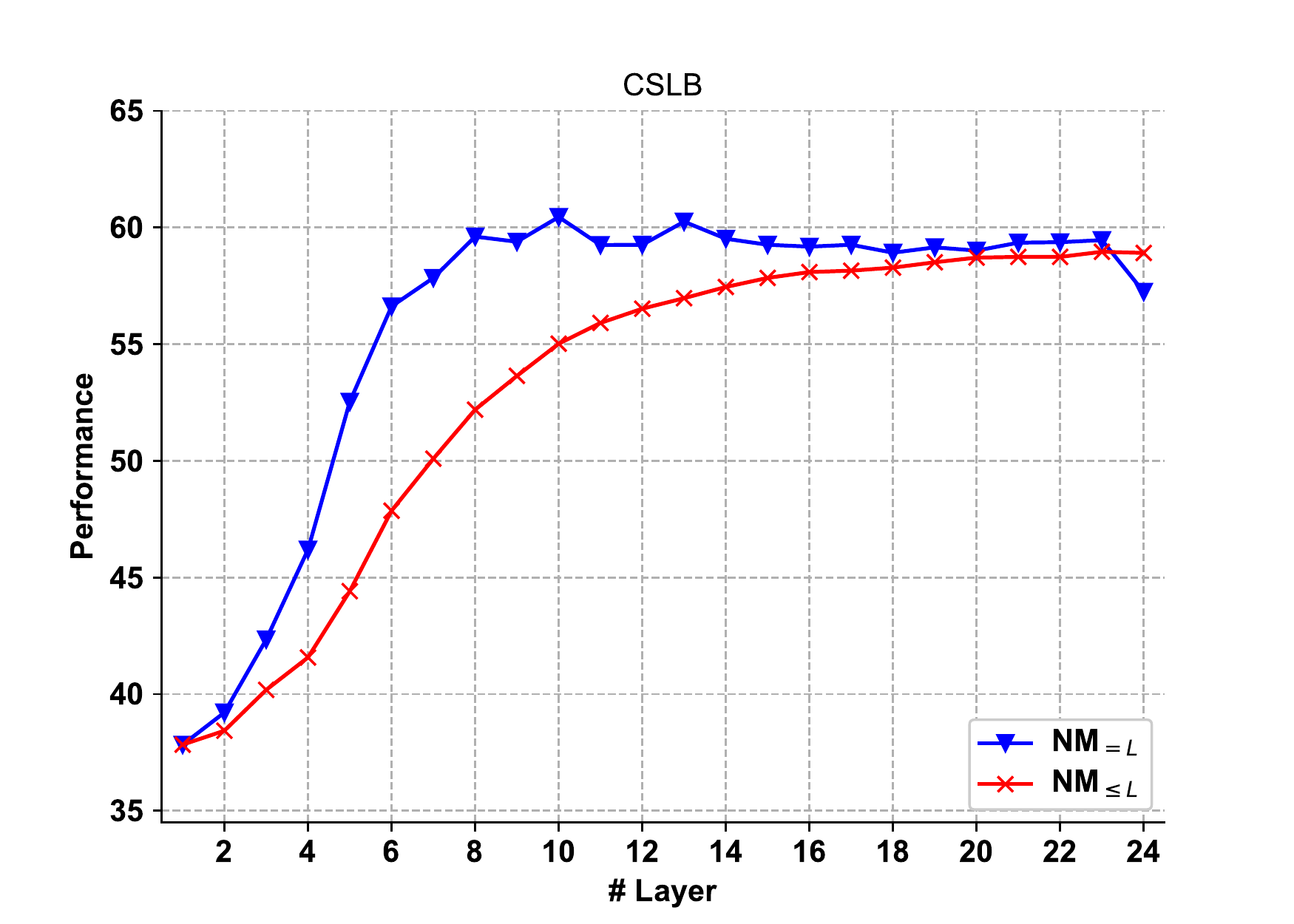}
\caption{Layer-wise performance on CSLB (RoBERTa).}
\label{figLayerPlot7}
\end{figure}

\begin{figure}[p]
\centering
\includegraphics[width=200pt]{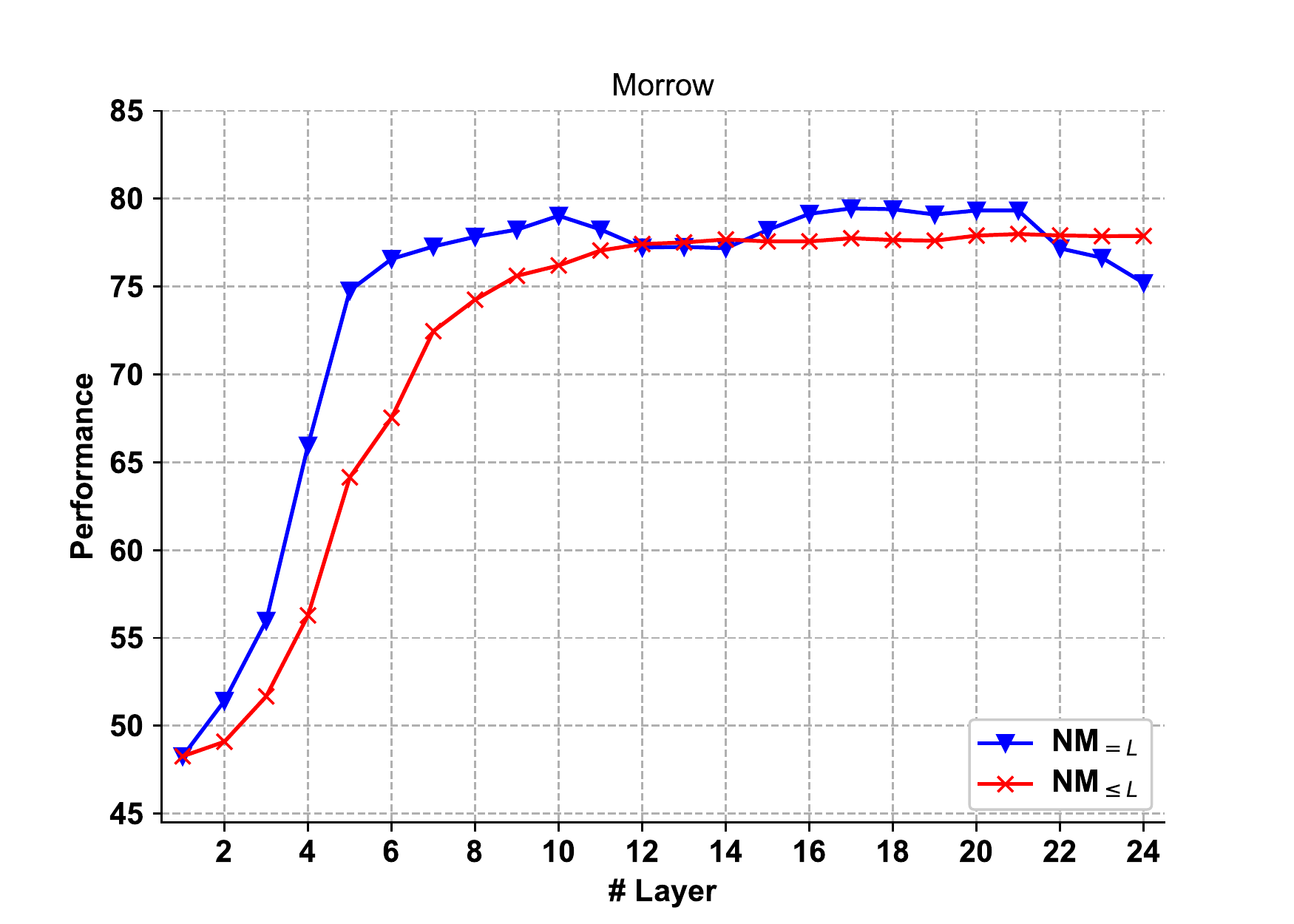}
\caption{Layer-wise performance on Morrow (RoBERTa).}
\label{figLayerPlot8}
\end{figure}

\begin{figure}[p]
\centering
\includegraphics[width=200pt]{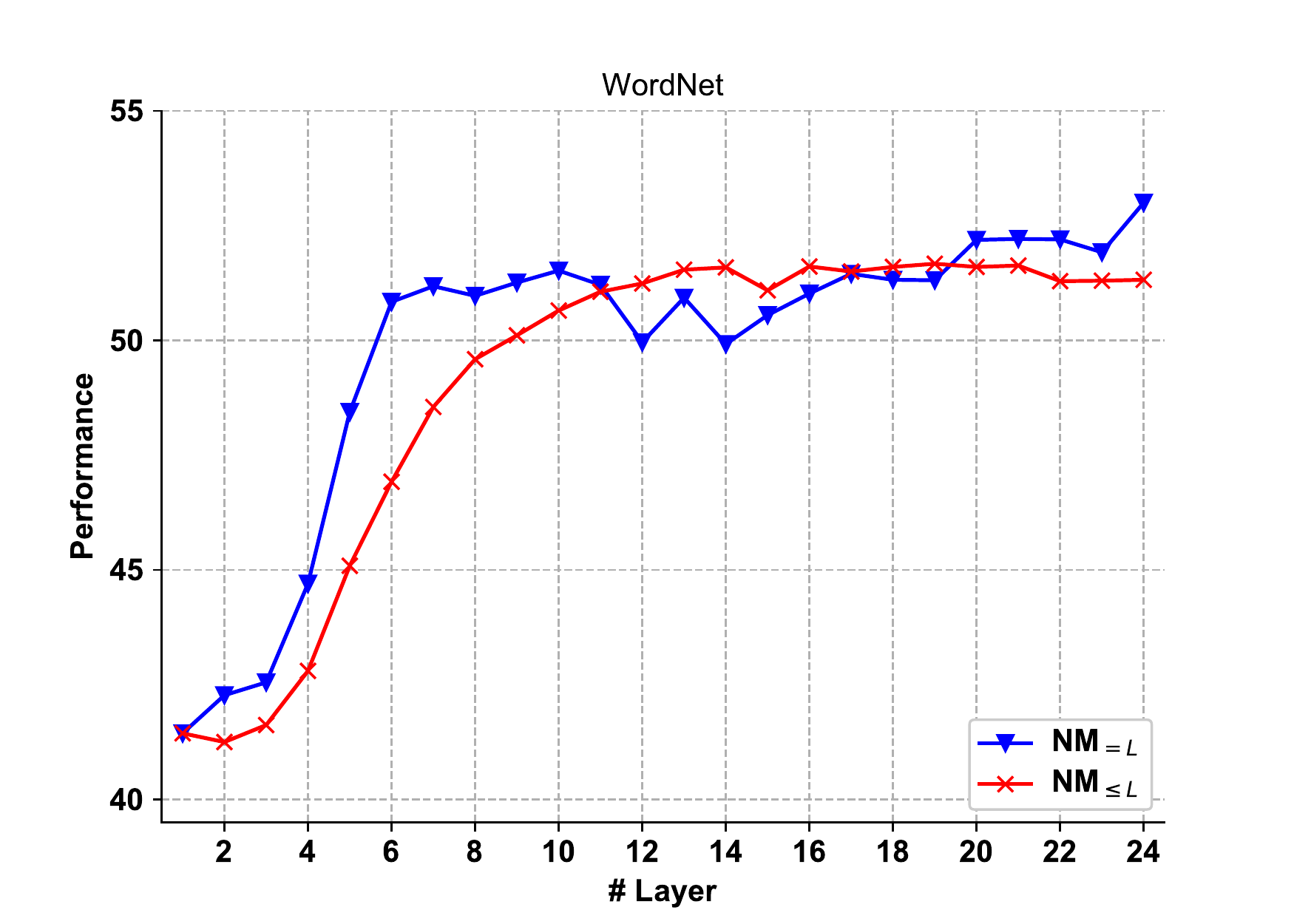}
\caption{Layer-wise performance on WordNet supersenses (RoBERTa).}
\label{figLayerPlot9}
\end{figure}

\begin{figure}[p]
\centering
\includegraphics[width=200pt]{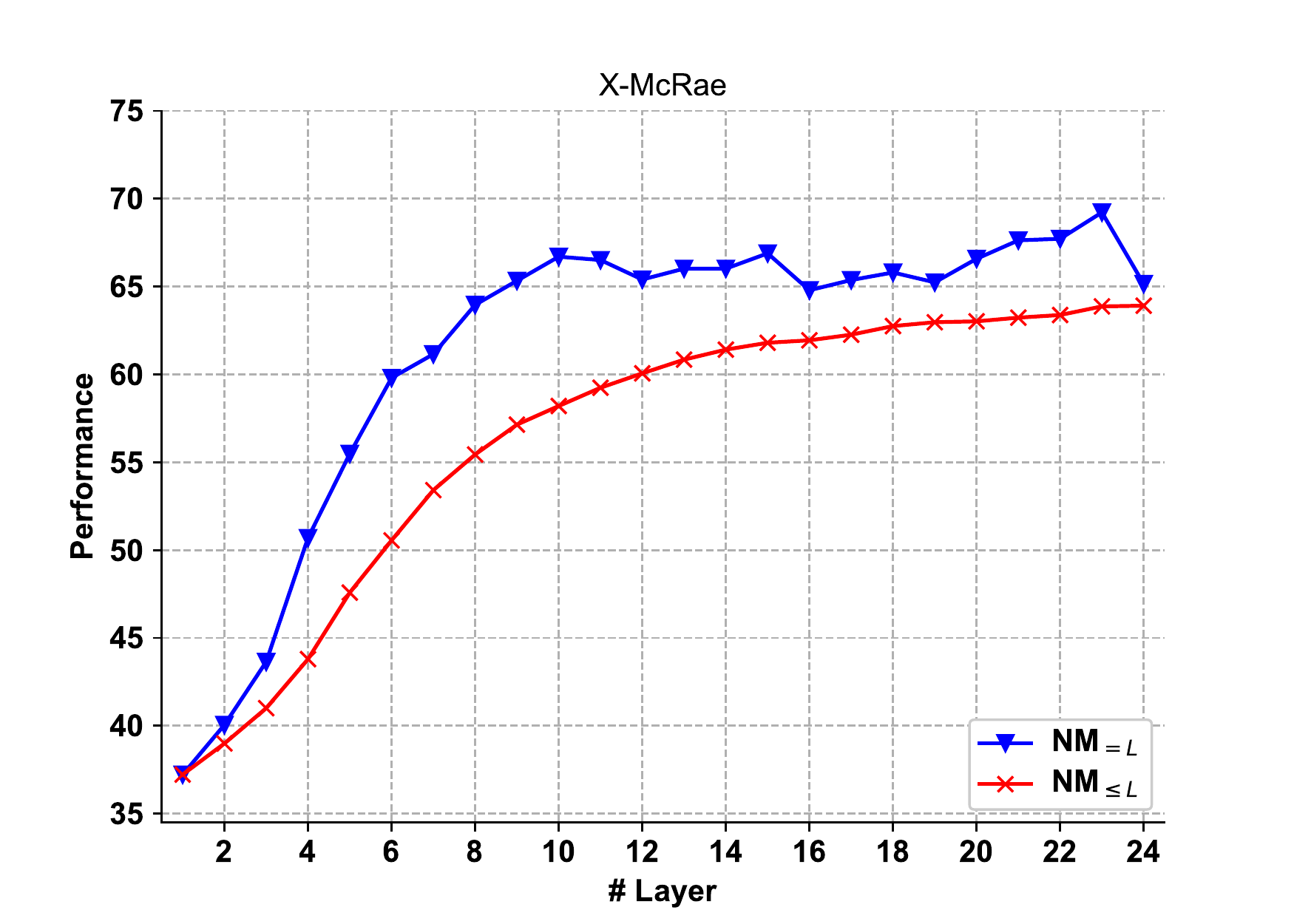}
\caption{Layer-wise performance on X-McRae (RoBERTa).}
\label{figLayerPlotLast}
\end{figure}

\noindent\textbf{Layerwise performance.} A detailed analysis of the performance of individual layers is provided in Tables \ref{each_svm_linear_bert_dev}--\ref{klayer_svm_linear_roberta_test}. In particular, Tables \ref{each_svm_linear_bert_dev} and \ref{each_svm_linear_bert_test} shows the results of NM$_{=L}$ for BERT on the tuning and test sets respectively. Tables \ref{each_svm_linear_roberta_dev} and \ref{each_svm_linear_roberta_test} show these results for RoBERTa. Tables \ref{klayer_svm_linear_bert_dev} and \ref{klayer_svm_linear_bert_test} show the results of NM$_{\leq L}$ for BERT, while Tables \ref{klayer_svm_linear_roberta_dev} and \ref{klayer_svm_linear_roberta_test} show these results for RoBERTa. A graphical presentation of the layer-wise performance on the test set is provided in Figures \ref{figLayerPlot1}--\ref{figLayerPlotLast}.

\begin{figure}[p]
\centering
\includegraphics[width=165pt]{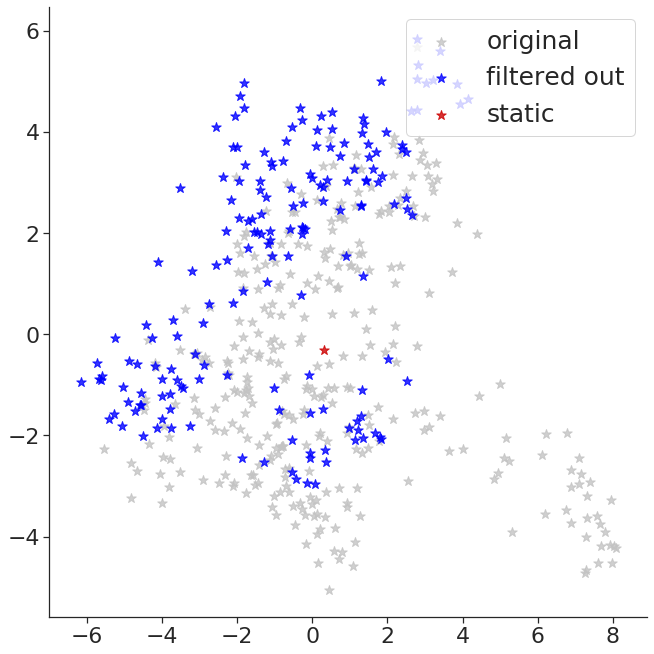}
\caption{Plots of the mention vectors for \emph{ash}, showing the mention vectors which are removed by the filtering strategy (blue) and the corresponding static vector (red).}
\label{figPlot1}
\end{figure}

\begin{figure}[p]
\centering
\includegraphics[width=165pt]{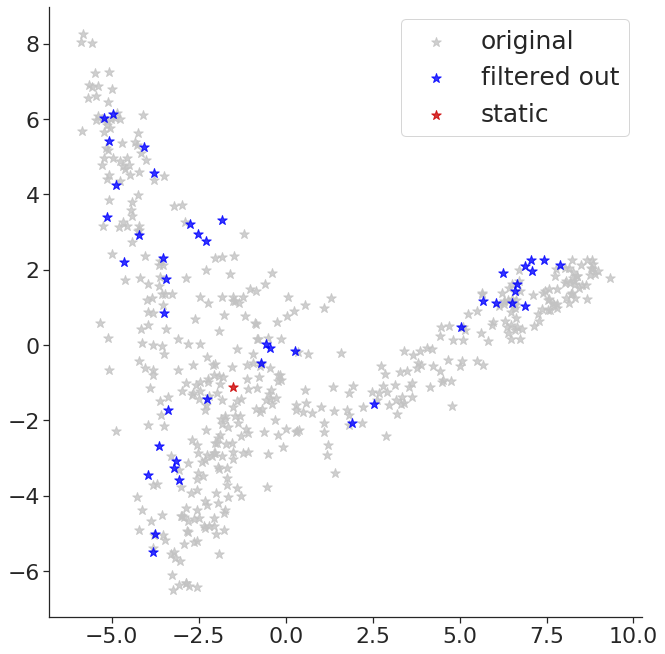}
\caption{Plots of the mention vectors for \emph{birch}, showing the mention vectors which are removed by the filtering strategy (blue) and the corresponding static vector (red).}
\label{figPlot2}
\end{figure}

\begin{figure}[p]
\centering
\includegraphics[width=165pt]{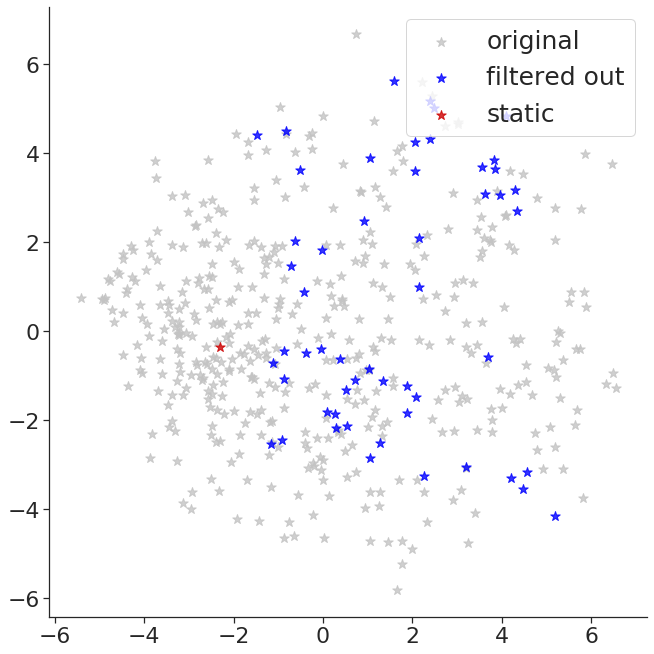}
\caption{Plots of the mention vectors for \emph{bookcase}, showing the mention vectors which are removed by the filtering strategy (blue) and the corresponding static vector (red).}
\label{figPlot3}
\end{figure}

\begin{figure}[p]
\centering
\includegraphics[width=165pt]{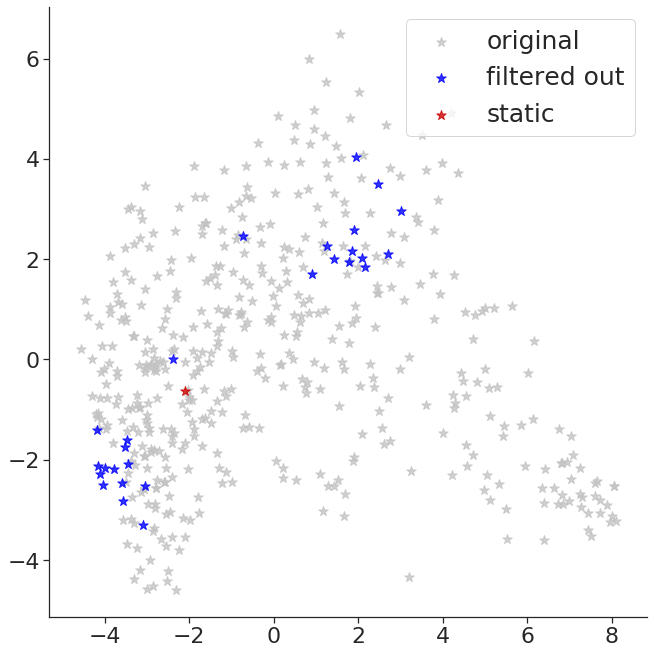}
\caption{Plots of the mention vectors for \emph{cicada}, showing the mention vectors which are removed by the filtering strategy (blue) and the corresponding static vector (red).}
\label{figPlot4}
\end{figure}

\begin{figure}[p]
\centering
\includegraphics[width=165pt]{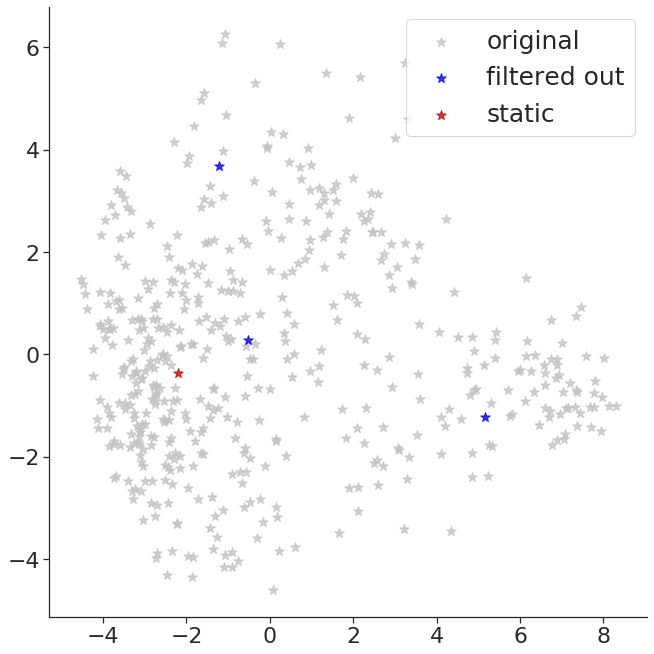}
\caption{Plots of the mention vectors for \emph{cockroach}, showing the mention vectors which are removed by the filtering strategy (blue) and the corresponding static vector (red).}
\label{figPlot5}
\end{figure}

\begin{figure}[p]
\centering
\includegraphics[width=165pt]{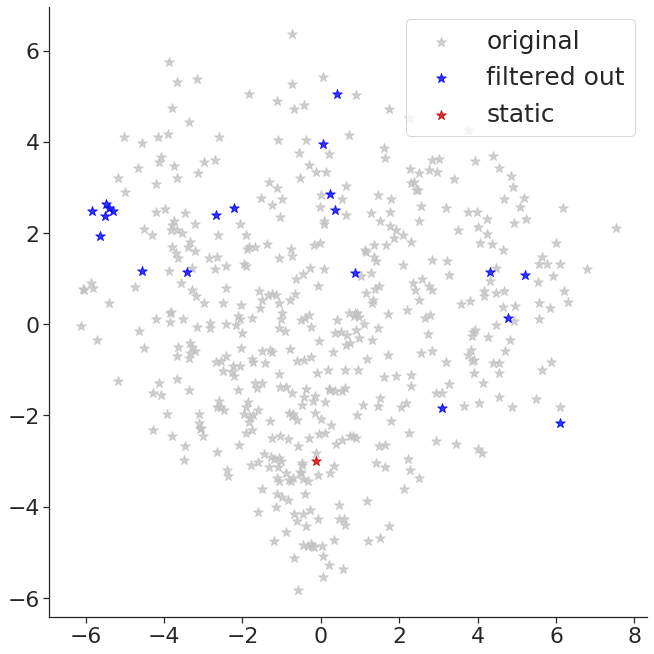}
\caption{Plots of the mention vectors for \emph{cupboard}, showing the mention vectors which are removed by the filtering strategy (blue) and the corresponding static vector (red).}
\label{figPlot6}
\end{figure}

\begin{figure}[p]
\centering
\includegraphics[width=165pt]{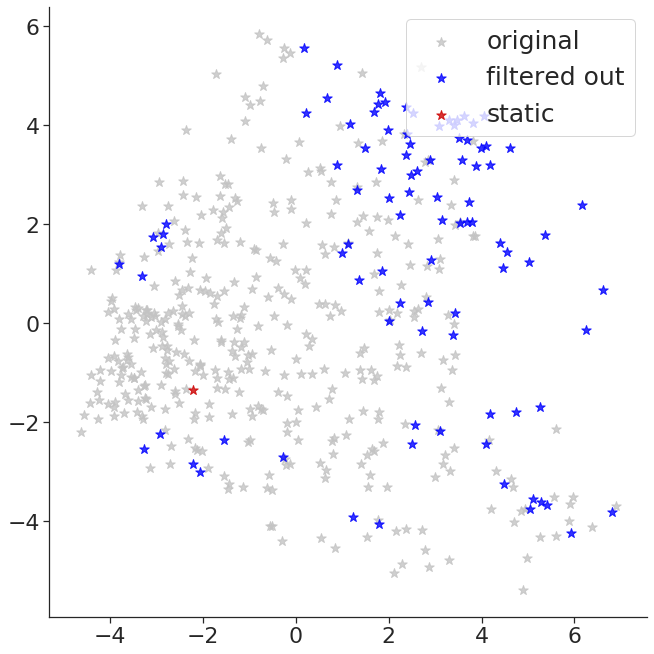}
\caption{Plots of the mention vectors for \emph{fridge}, showing the mention vectors which are removed by the filtering strategy (blue) and the corresponding static vector (red).}
\label{figPlot7}
\end{figure}

\begin{figure}[p]
\centering
\includegraphics[width=165pt]{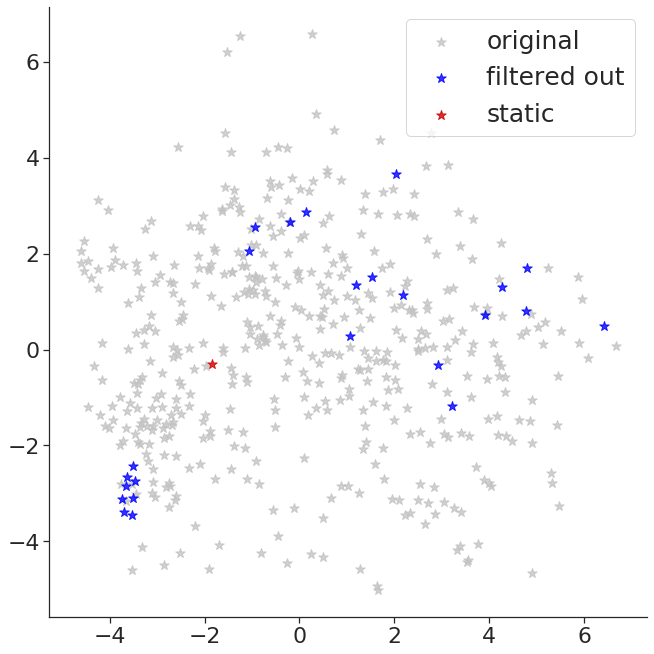}
\caption{Plots of the mention vectors for \emph{maggot}, showing the mention vectors which are removed by the filtering strategy (blue) and the corresponding static vector (red).}
\label{figPlot8}
\end{figure}

\begin{figure}[p]
\centering
\includegraphics[width=165pt]{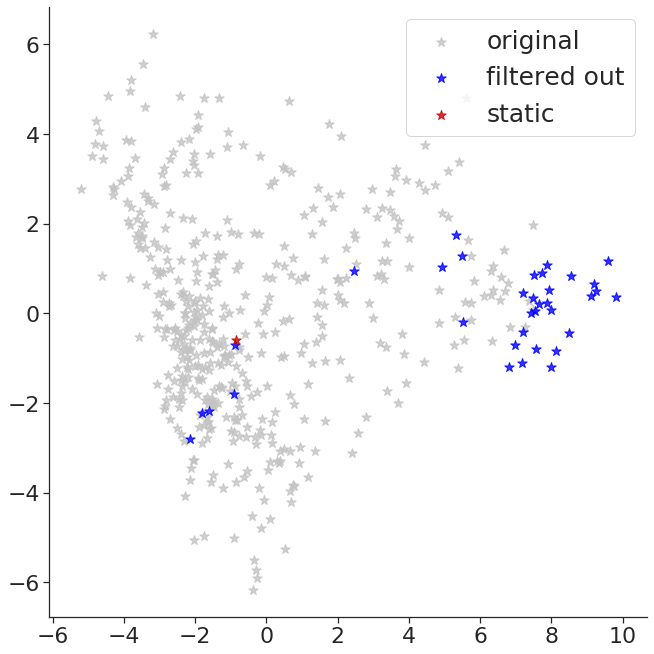}
\caption{Plots of the mention vectors for \emph{mantis}, showing the mention vectors which are removed by the filtering strategy (blue) and the corresponding static vector (red).}
\label{figPlot9}
\end{figure}

\begin{figure}[p]
\centering
\includegraphics[width=165pt]{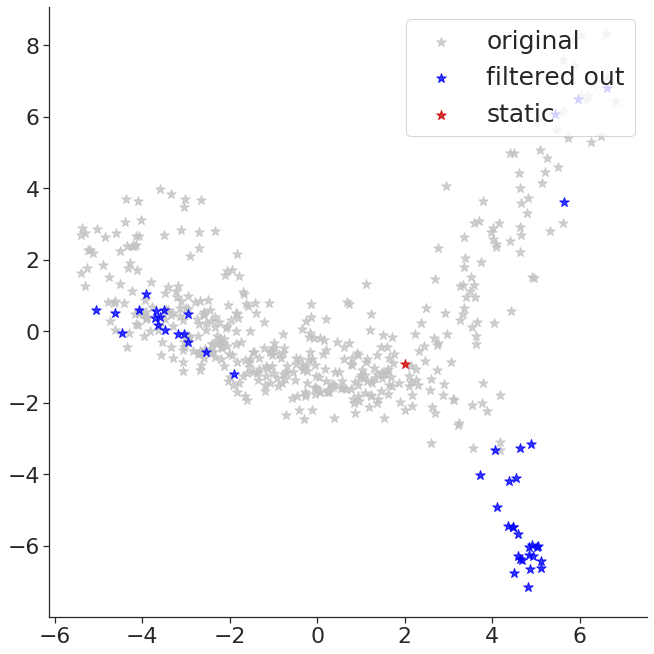}
\caption{Plots of the mention vectors for \emph{mosquito}, showing the mention vectors which are removed by the filtering strategy (blue) and the corresponding static vector (red).}
\label{figPlot10}
\end{figure}

\begin{figure}[p]
\centering
\includegraphics[width=165pt]{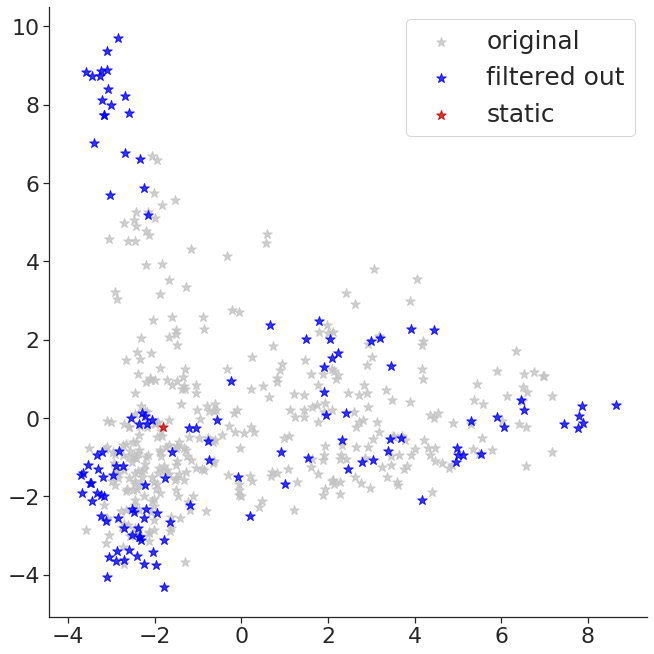}
\caption{Plots of the mention vectors for \emph{palm}, showing the mention vectors which are removed by the filtering strategy (blue) and the corresponding static vector (red).}
\label{figPlot11}
\end{figure}

\begin{figure}[p]
\centering
\includegraphics[width=165pt]{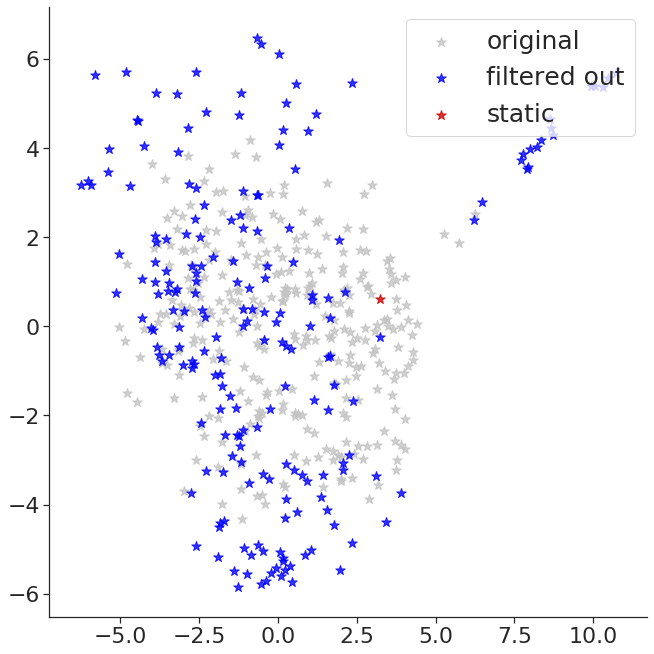}
\caption{Plots of the mention vectors for \emph{sapling}, showing the mention vectors which are removed by the filtering strategy (blue) and the corresponding static vector (red).}
\label{figPlot12}
\end{figure}

\begin{figure}[p]
\centering
\includegraphics[width=165pt]{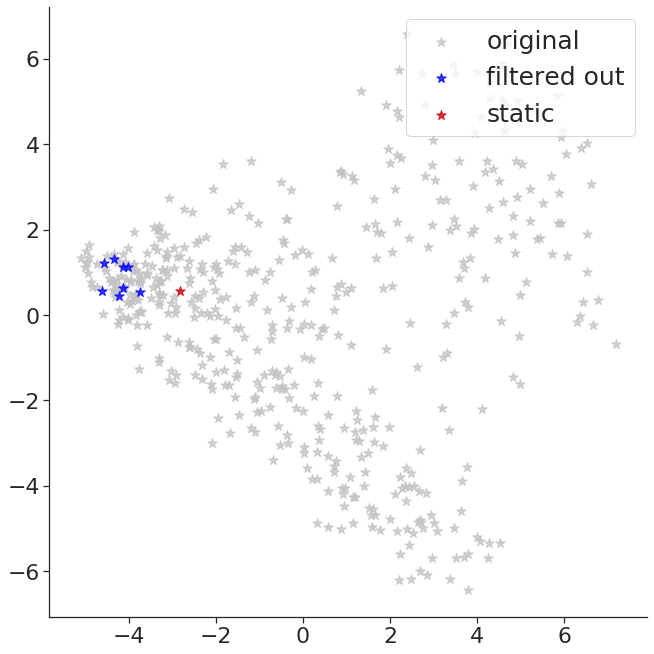}
\caption{Plots of the mention vectors for \emph{sofa}, showing the mention vectors which are removed by the filtering strategy (blue) and the corresponding static vector (red).}
\label{figPlot13}
\end{figure}

\begin{figure}[p]
\centering
\includegraphics[width=165pt]{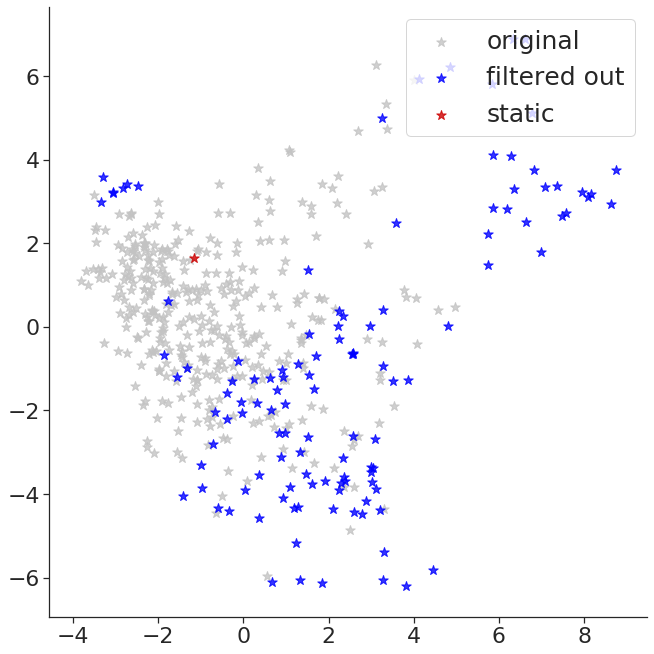}
\caption{Plots of the mention vectors for \emph{typewriter}, showing the mention vectors which are removed by the filtering strategy (blue) and the corresponding static vector (red).}
\label{figPlot14}
\end{figure}

\begin{figure}[p]
\centering
\includegraphics[width=165pt]{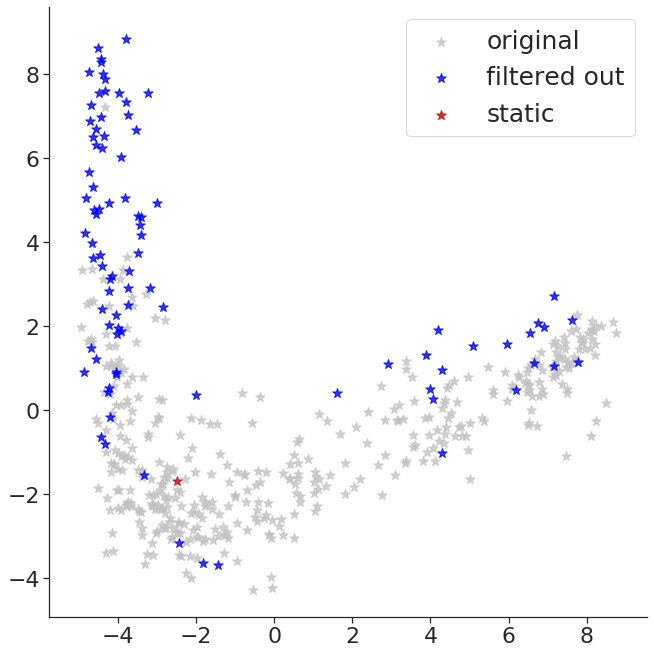}
\caption{Plots of the mention vectors for \emph{willow}, showing the mention vectors which are removed by the filtering strategy (blue) and the corresponding static vector (red).}
\label{figPlot15}
\end{figure}

\medskip
\noindent\textbf{Qualitative analysis.} Figures \ref{figPlot1}--\ref{figPlot15} show which mention vectors were filtered using the proposed strategy, for a number of different target words.

%******************************************************************************************************************************************************

\begin{table*}[p]
\centering
\footnotesize
\begin{tabular}{
@{}c
cc
cc
cc
cc
cc
}
\toprule  
\multirow{2}{0.3in}{\textbf{Layer}} 
& \multicolumn{2}{c}{\textbf{X-McRae}}
& \multicolumn{2}{c}{\textbf{CSLB}}
& \multicolumn{2}{c}{\textbf{Morrow}}
& \multicolumn{2}{c}{\textbf{WordNet}}
& \multicolumn{2}{c}{\textbf{BabelNet}}\\
\cmidrule(lr){2-3}
\cmidrule(lr){4-5}
\cmidrule(lr){6-7}
\cmidrule(lr){8-9}
\cmidrule(lr){10-11}
& MAP & F1        
& MAP & F1 
& MAP & F1 
& MAP & F1 
& MAP & F1\\
 \midrule
        1 & 56.8 & 60.0 & 58.7 & 63.4 & 57.5 & 62.6 & 35.1 & 41.0 & 32.3 & 39.0 \\ %\hline
        2 & 56.7 & 59.9 & 59.3 & 64.4 & 60.3 & 65.8 & 37.1 & 41.2 & 34.6 & 40.8 \\ %\hline
        3 & 59.5 & 61.3 & 60.0 & 65.1 & 64.7 & 69.2 & 38.9 & 43.5 & 34.9 & 41.8 \\ %\hline
        4 & 60.4 & 62.4 & 61.1 & 65.9 & 66.2 & 68.3 & 40.0 & 44.5 & 34.9 & 42.7 \\ %\hline
        5 & 63.0 & 64.3 & 61.9 & 66.6 & 70.8 & 71.2 & 41.9 & 45.8 & 35.7 & 43.0 \\ %\hline
        6 & 65.5 & 66.7 & 63.1 & 68.0 & 72.2 & 74.6 & 44.6 & 48.0 & 37.2 & 43.6 \\ %\hline
        7 & 65.8 & 67.8 & 63.0 & 68.1 & 73.3 & 74.1 & 45.2 & 48.2 & 38.2 & 44.1 \\ %\hline
        8 & 65.7 & 67.3 & 62.1 & 67.5 & 74.2 & 74.0 & 45.5 & 48.9 & 38.9 & 44.8 \\ %\hline
        9 & 67.1 & 68.4 & 62.6 & 67.9 & 74.7 & 74.5 & 45.1 & 48.7 & 38.8 & 44.4 \\ %\hline
        10 & 66.6 & 68.9 & 61.6 & 66.8 & 72.8 & 74.0 & 43.6 & 47.0 & 38.0 & 45.2 \\ %\hline
        11 & 65.6 & 67.3 & 61.2 & 66.5 & 72.3 & 72.0 & 43.3 & 47.3 & 37.6 & 44.0 \\ %\hline
        12 & 66.0 & 67.0 & 60.9 & 66.4 & 71.2 & 71.1 & 43.1 & 46.8 & 36.4 & 43.4 \\ %\hline
        13 & 65.8 & 67.0 & 60.5 & 65.9 & 70.3 & 70.3 & 42.6 & 46.8 & 37.3 & 43.6 \\ %\hline
        14 & 67.1 & 68.2 & 60.4 & 65.8 & 71.2 & 70.8 & 42.6 & 46.8 & 37.2 & 43.4 \\ %\hline
        15 & 67.3 & 67.9 & 61.1 & 66.5 & 72.5 & 72.1 & 42.9 & 47.0 & 37.9 & 43.8 \\ %\hline
        16 & 67.2 & 67.9 & 61.3 & 66.5 & 72.4 & 72.0 & 43.0 & 47.6 & 37.5 & 43.6 \\ %\hline
        17 & 67.3 & 67.7 & 60.7 & 66.3 & 71.5 & 71.3 & 43.3 & 47.3 & 36.9 & 43.3 \\ %\hline
        18 & 67.1 & 68.3 & 61.2 & 67.0 & 72.8 & 73.4 & 44.3 & 48.0 & 36.9 & 43.5 \\ %\hline
        19 & 67.5 & 68.8 & 61.3 & 66.7 & 73.0 & 73.2 & 44.5 & 48.1 & 38.1 & 43.7 \\ %\hline
        20 & 67.8 & 69.2 & 61.5 & 66.7 & 71.2 & 71.2 & 44.1 & 47.7 & 37.3 & 42.9 \\ %\hline
        21 & 68.3 & 70.0 & 62.1 & 67.3 & 72.8 & 73.4 & 44.8 & 48.3 & 38.5 & 43.1 \\ %\hline
        22 & 68.9 & 70.3 & 62.5 & 67.4 & 73.5 & 72.2 & 44.4 & 47.3 & 38.7 & 43.5 \\ %\hline
        23 & 68.6 & 69.4 & 62.0 & 66.8 & 70.2 & 70.7 & 43.6 & 46.4 & 37.1 & 42.2 \\ %\hline
        24 & 67.8 & 68.3 & 61.7 & 66.3 & 70.5 & 69.9 & 41.6 & 45.2 & 37.2 & 42.4 \\ %\hline
\bottomrule  
\end{tabular}
\caption{Results ($\%$) on \textbf{validation set} for classification tasks in each layer in terms of MAP and F1 scores, using mention vectors extracted from each layer of BERT.}
\label{each_svm_linear_bert_dev}
\end{table*}

% bert + test
\begin{table*}[p]
\centering
\footnotesize
\begin{tabular}{
@{}c
cc
cc
cc
cc
cc
}
\toprule  
\multirow{2}{0.3in}{\textbf{Layer}} 
& \multicolumn{2}{c}{\textbf{X-McRae}}
& \multicolumn{2}{c}{\textbf{CSLB}}
& \multicolumn{2}{c}{\textbf{Morrow}}
& \multicolumn{2}{c}{\textbf{WordNet}}
& \multicolumn{2}{c}{\textbf{BabelNet}}\\
\cmidrule(lr){2-3}
\cmidrule(lr){4-5}
\cmidrule(lr){6-7}
\cmidrule(lr){8-9}
\cmidrule(lr){10-11}
& MAP & F1        
& MAP & F1 
& MAP & F1 
& MAP & F1 
& MAP & F1\\
 \midrule
1 & 64.5 & 55.7 & 58.6 & 42.8 & 62.1 & 50.1 & 33.7 & 36.3 & 33.8 & 31.6 \\ %\hline
2 & 64.2 & 55.4 & 59.3 & 44.2 & 63.5 & 53.4 & 37.0 & 38.5 & 35.8 & 35.0 \\ %\hline
3 & 66.7 & 58.1 & 59.7 & 45.5 & 67.9 & 55.7 & 37.5 & 40.0 & 37.4 & 35.4 \\ %\hline
4 & 68.1 & 57.7 & 61.0 & 46.1 & 68.7 & 57.8 & 39.1 & 39.8 & 38.6 & 36.9 \\ %\hline
5 & 69.3 & 59.4 & 61.9 & 47.6 & 72.5 & 58.4 & 40.5 & 40.3 & 40.2 & 38.7 \\ %\hline
6 & 70.9 & 59.3 & 62.7 & 49.7 & 74.2 & 58.5 & 43.9 & 42.2 & 42.2 & 41.4 \\ %\hline
7 & 70.7 & 58.6 & 63.5 & 49.9 & 76.9 & 62.8 & 45.0 & 43.3 & 42.8 & 42.0 \\ %\hline
8 & 70.1 & 61.7 & 62.9 & 49.6 & 77.5 & 62.4 & 46.7 & 44.6 & 43.4 & 41.2 \\ %\hline
9 & 70.9 & 61.2 & 63.1 & 49.6 & 78.5 & 64.1 & 47.1 & 44.4 & 42.8 & 40.7 \\ %\hline
10 & 71.2 & 60.3 & 62.5 & 49.1 & 77.6 & 64.2 & 45.3 & 44.4 & 43.0 & 39.3 \\ %\hline
11 & 70.4 & 60.9 & 61.8 & 49.1 & 76.4 & 63.6 & 45.0 & 43.4 & 41.6 & 39.9 \\ %\hline
12 & 70.8 & 60.7 & 61.8 & 49.3 & 75.8 & 61.6 & 43.7 & 43.1 & 40.8 & 40.8 \\ %\hline
13 & 71.0 & 60.5 & 61.1 & 48.7 & 73.3 & 58.9 & 44.1 & 42.4 & 40.9 & 39.2 \\ %\hline
14 & 71.1 & 58.6 & 61.0 & 48.7 & 73.4 & 59.0 & 44.5 & 42.6 & 40.7 & 39.0 \\ %\hline
15 & 71.6 & 58.4 & 62.4 & 49.7 & 76.0 & 62.9 & 44.7 & 44.2 & 42.0 & 40.2 \\ %\hline
16 & 71.2 & 59.7 & 62.4 & 48.1 & 76.8 & 61.5 & 44.7 & 44.4 & 42.4 & 40.3 \\ %\hline
17 & 71.7 & 60.7 & 62.2 & 47.5 & 75.9 & 64.8 & 44.9 & 44.9 & 41.4 & 37.7 \\ %\hline
18 & 72.2 & 61.8 & 62.8 & 47.8 & 76.9 & 64.2 & 45.6 & 44.1 & 42.2 & 38.8 \\ %\hline
19 & 71.8 & 63.1 & 63.0 & 49.3 & 77.5 & 64.2 & 45.6 & 43.8 & 43.3 & 40.8 \\ %\hline
20 & 71.9 & 63.1 & 63.1 & 48.1 & 77.7 & 63.1 & 44.8 & 43.2 & 42.5 & 37.9 \\ %\hline
21 & 71.0 & 62.5 & 63.5 & 49.7 & 78.0 & 67.5 & 45.9 & 44.8 & 43.3 & 40.2 \\ %\hline
22 & 70.4 & 60.4 & 63.1 & 49.3 & 78.4 & 66.5 & 46.4 & 44.7 & 44.3 & 38.1 \\ %\hline
23 & 70.1 & 61.6 & 62.1 & 48.6 & 77.0 & 65.7 & 44.4 & 43.3 & 42.5 & 39.0 \\ %\hline
24 & 69.4 & 61.5 & 61.4 & 48.1 & 75.9 & 65.7 & 42.5 & 41.6 & 41.0 & 36.8 \\ %\hline
\bottomrule  
\end{tabular}
\caption{Results ($\%$) on \textbf{test set} for classification tasks in each layer in terms of MAP and F1 scores, using  mention vectors extracted from each layer of BERT.}
\label{each_svm_linear_bert_test}
\end{table*}

% roberta + dev
\begin{table*}[p]
\centering
\footnotesize
\begin{tabular}{
@{}c
cc
cc
cc
cc
cc
}
\toprule  
\multirow{2}{0.3in}{\textbf{Layer}} 
& \multicolumn{2}{c}{\textbf{X-McRae}}
& \multicolumn{2}{c}{\textbf{CSLB}}
& \multicolumn{2}{c}{\textbf{Morrow}}
& \multicolumn{2}{c}{\textbf{WordNet}}
& \multicolumn{2}{c}{\textbf{BabelNet}}\\
\cmidrule(lr){2-3}
\cmidrule(lr){4-5}
\cmidrule(lr){6-7}
\cmidrule(lr){8-9}
\cmidrule(lr){10-11}
& MAP & F1        
& MAP & F1 
& MAP & F1 
& MAP & F1 
& MAP & F1\\
 \midrule
        1 & 37.2 & 45.7 & 36.8 & 46.9 & 43.2 & 51.1 & 39.8 & 45.2 & 35.0 & 41.4 \\ %\hline
        2 & 38.5 & 46.0 & 38.5 & 48.4 & 48.2 & 53.4 & 41.3 & 46.2 & 34.6 & 41.7 \\ %\hline
        3 & 42.0 & 48.7 & 40.8 & 50.5 & 54.6 & 57.2 & 42.7 & 46.3 & 35.7 & 42.7 \\ %\hline
        4 & 47.8 & 52.2 & 44.6 & 53.1 & 57.8 & 60.9 & 43.1 & 47.3 & 35.9 & 43.6 \\ %\hline
        5 & 53.8 & 57.8 & 50.4 & 57.6 & 64.4 & 64.5 & 45.6 & 49.4 & 37.9 & 45.2 \\ %\hline
        6 & 59.0 & 61.7 & 55.5 & 61.8 & 70.7 & 70.6 & 48.5 & 51.8 & 40.5 & 46.8 \\ %\hline
        7 & 61.3 & 63.3 & 57.3 & 63.2 & 70.7 & 71.5 & 49.8 & 52.2 & 41.6 & 47.2 \\ %\hline
        8 & 62.8 & 65.6 & 58.8 & 64.8 & 71.7 & 72.9 & 49.5 & 52.9 & 42.3 & 48.1 \\ %\hline
        9 & 64.9 & 66.3 & 59.0 & 64.7 & 72.2 & 73.3 & 49.5 & 53.1 & 42.0 & 47.5 \\ %\hline
        10 & 64.6 & 67.0 & 60.1 & 65.4 & 73.7 & 73.3 & 49.3 & 52.5 & 42.3 & 48.6 \\ %\hline
        11 & 64.5 & 67.0 & 59.7 & 65.3 & 72.3 & 71.1 & 49.1 & 52.8 & 42.3 & 48.5 \\ %\hline
        12 & 63.7 & 66.0 & 59.4 & 65.4 & 72.8 & 70.6 & 49.7 & 52.9 & 41.9 & 48.3 \\ %\hline
        13 & 64.5 & 66.4 & 59.9 & 65.8 & 73.3 & 71.6 & 49.9 & 53.2 & 41.4 & 48.3 \\ %\hline
        14 & 65.2 & 67.0 & 59.7 & 65.5 & 71.3 & 69.6 & 49.5 & 52.1 & 40.9 & 47.6 \\ %\hline
        15 & 64.7 & 66.9 & 59.4 & 65.4 & 72.5 & 70.8 & 49.3 & 51.9 & 41.3 & 47.8 \\ %\hline
        16 & 65.6 & 66.5 & 59.3 & 64.9 & 72.3 & 70.3 & 49.2 & 51.8 & 41.8 & 47.8 \\ %\hline
        17 & 65.9 & 67.2 & 59.0 & 64.8 & 72.0 & 70.6 & 48.8 & 51.7 & 41.8 & 48.6 \\ %\hline
        18 & 65.4 & 66.3 & 58.7 & 64.7 & 71.9 & 71.3 & 49.1 & 51.6 & 41.5 & 48.6 \\ %\hline
        19 & 66.9 & 67.5 & 59.5 & 65.2 & 72.5 & 70.3 & 49.8 & 52.2 & 41.4 & 48.1 \\ %\hline
        20 & 66.3 & 66.3 & 59.0 & 64.6 & 72.1 & 71.1 & 49.7 & 52.2 & 40.9 & 47.6 \\ %\hline
        21 & 66.0 & 67.6 & 59.3 & 65.0 & 72.0 & 70.7 & 49.8 & 52.4 & 40.3 & 46.9 \\ %\hline
        22 & 66.2 & 67.1 & 59.5 & 64.7 & 71.7 & 70.8 & 49.3 & 52.7 & 40.3 & 47.0 \\ %\hline
        23 & 65.7 & 66.6 & 58.9 & 64.5 & 72.3 & 70.8 & 48.7 & 52.7 & 40.6 & 47.2 \\ %\hline
        24 & 60.7 & 64.1 & 56.5 & 62.4 & 70.0 & 69.7 & 50.5 & 53.6 & 41.4 & 48.2 \\ %\hline
\bottomrule  
\end{tabular}
\caption{Results ($\%$) on \textbf{validation set} for classification tasks in each layer in terms of MAP and F1 scores, using mention vectors extracted from each layer of RoBERTa.}
\label{each_svm_linear_roberta_dev}
\end{table*}

% roberta + test
\begin{table*}[p]
\centering
\footnotesize
\begin{tabular}{
@{}c
cc
cc
cc
cc
cc
}
\toprule  
\multirow{2}{0.3in}{\textbf{Layer}} 
& \multicolumn{2}{c}{\textbf{X-McRae}}
& \multicolumn{2}{c}{\textbf{CSLB}}
& \multicolumn{2}{c}{\textbf{Morrow}}
& \multicolumn{2}{c}{\textbf{WordNet}}
& \multicolumn{2}{c}{\textbf{BabelNet}}\\
\cmidrule(lr){2-3}
\cmidrule(lr){4-5}
\cmidrule(lr){6-7}
\cmidrule(lr){8-9}
\cmidrule(lr){10-11}
& MAP & F1        
& MAP & F1 
& MAP & F1 
& MAP & F1 
& MAP & F1\\
 \midrule
        1 & 37.2 & 32.4 & 37.8 & 27.7 & 48.3 & 43.9 & 41.4 & 41.3 & 33.2 & 30.5 \\ %\hline
        2 & 40.1 & 35.9 & 39.2 & 28.5 & 51.4 & 41.9 & 42.3 & 42.0 & 35.0 & 34.4 \\ %\hline
        3 & 43.6 & 39.6 & 42.3 & 29.7 & 56.0 & 42.3 & 42.6 & 41.7 & 36.2 & 35.0 \\ %\hline
        4 & 50.7 & 44.6 & 46.2 & 33.4 & 66.0 & 58.8 & 44.7 & 43.6 & 37.0 & 35.2 \\ %\hline
        5 & 55.5 & 49.2 & 52.5 & 37.9 & 74.8 & 64.2 & 48.4 & 45.9 & 39.9 & 38.8 \\ %\hline
        6 & 59.8 & 50.0 & 56.6 & 43.0 & 76.6 & 61.3 & 50.8 & 48.0 & 42.8 & 39.4 \\ %\hline
        7 & 61.2 & 51.0 & 57.8 & 43.3 & 77.3 & 63.6 & 51.2 & 49.0 & 43.2 & 42.7 \\ %\hline
        8 & 64.0 & 54.9 & 59.6 & 45.2 & 77.8 & 65.4 & 51.0 & 49.3 & 43.4 & 40.3 \\ %\hline
        9 & 65.3 & 54.6 & 59.4 & 44.7 & 78.2 & 62.2 & 51.3 & 50.1 & 43.7 & 39.6 \\ %\hline
        10 & 66.7 & 56.0 & 60.5 & 45.9 & 79.0 & 67.5 & 51.5 & 49.2 & 43.3 & 41.0 \\ %\hline
        11 & 66.5 & 57.5 & 59.3 & 44.5 & 78.3 & 69.3 & 51.2 & 49.0 & 42.7 & 39.6 \\ %\hline
        12 & 65.4 & 56.9 & 59.3 & 45.1 & 77.2 & 67.0 & 50.0 & 50.1 & 42.2 & 40.7 \\ %\hline
        13 & 66.0 & 57.0 & 60.3 & 45.6 & 77.3 & 68.4 & 50.9 & 50.2 & 42.3 & 39.0 \\ %\hline
        14 & 66.0 & 55.9 & 59.5 & 45.2 & 77.2 & 68.2 & 49.9 & 49.2 & 41.6 & 41.0 \\ %\hline
        15 & 66.9 & 56.2 & 59.3 & 44.7 & 78.2 & 67.4 & 50.6 & 50.0 & 41.9 & 38.8 \\ %\hline
        16 & 64.8 & 57.2 & 59.2 & 44.8 & 79.1 & 67.5 & 51.0 & 49.2 & 42.2 & 39.1 \\ %\hline
        17 & 65.4 & 57.2 & 59.3 & 44.2 & 79.5 & 69.1 & 51.5 & 50.6 & 42.2 & 39.8 \\ %\hline
        18 & 65.8 & 57.0 & 58.9 & 43.8 & 79.4 & 67.9 & 51.3 & 49.2 & 42.4 & 40.4 \\ %\hline
        19 & 65.2 & 56.5 & 59.2 & 43.6 & 79.1 & 66.5 & 51.3 & 50.4 & 42.7 & 41.6 \\ %\hline
        20 & 66.6 & 55.4 & 59.0 & 44.0 & 79.3 & 69.5 & 52.2 & 51.4 & 42.5 & 39.8 \\ %\hline
        21 & 67.6 & 55.9 & 59.4 & 43.9 & 79.3 & 68.7 & 52.2 & 50.7 & 43.3 & 40.6 \\ %\hline
        22 & 67.7 & 58.4 & 59.4 & 45.9 & 77.2 & 60.7 & 52.2 & 50.1 & 42.9 & 39.5 \\ %\hline
        23 & 69.2 & 57.6 & 59.5 & 45.8 & 76.6 & 63.5 & 51.9 & 51.0 & 43.3 & 40.3 \\ %\hline
        24 & 65.2 & 54.4 & 57.2 & 42.9 & 75.2 & 62.5 & 53.0 & 50.9 & 41.2 & 40.5 \\ %\hline
\bottomrule  
\end{tabular}
\caption{Results ($\%$) on \textbf{test set} for classification tasks in each layer in terms of MAP and F1 scores, using mention vectors extracted from each layer of RoBERTa.}
\label{each_svm_linear_roberta_test}
\end{table*}

% Nomask + avg of first k layers
% bert + dev
\begin{table*}[p]
\centering
\footnotesize
\begin{tabular}{
@{}c
cc
cc
cc
cc
cc
}
\toprule  
\multirow{2}{0.3in}{\textbf{Layer}} 
& \multicolumn{2}{c}{\textbf{X-McRae}}
& \multicolumn{2}{c}{\textbf{CSLB}}
& \multicolumn{2}{c}{\textbf{Morrow}}
& \multicolumn{2}{c}{\textbf{WordNet}}
& \multicolumn{2}{c}{\textbf{BabelNet}}\\
\cmidrule(lr){2-3}
\cmidrule(lr){4-5}
\cmidrule(lr){6-7}
\cmidrule(lr){8-9}
\cmidrule(lr){10-11}
& MAP & F1        
& MAP & F1 
& MAP & F1 
& MAP & F1 
& MAP & F1\\
 \midrule
        1 & 56.8 & 60.0 & 58.7 & 63.4 & 57.5 & 62.6 & 35.1 & 41.0 & 32.3 & 39.0 \\ %\hline
        2 & 57.0 & 59.5 & 59.0 & 64.1 & 59.3 & 65.0 & 36.5 & 41.6 & 33.7 & 40.3 \\ %\hline
        3 & 57.6 & 60.1 & 59.7 & 64.6 & 61.6 & 66.1 & 37.3 & 42.1 & 33.9 & 40.7 \\ %\hline
        4 & 58.7 & 60.9 & 60.0 & 64.8 & 63.4 & 67.5 & 37.9 & 42.2 & 34.3 & 41.2 \\ %\hline
        5 & 59.5 & 61.3 & 60.7 & 65.3 & 64.1 & 68.9 & 39.2 & 43.2 & 34.7 & 42.0 \\ %\hline
        6 & 60.8 & 62.5 & 61.5 & 66.1 & 66.4 & 69.8 & 40.9 & 44.8 & 35.3 & 42.5 \\ %\hline
        7 & 62.1 & 63.9 & 61.8 & 66.6 & 69.2 & 70.2 & 42.1 & 45.9 & 35.6 & 42.6 \\ %\hline
        8 & 63.1 & 64.8 & 62.2 & 67.0 & 70.8 & 71.6 & 43.2 & 46.6 & 36.1 & 43.1 \\ %\hline
        9 & 63.6 & 65.7 & 62.4 & 67.3 & 72.1 & 72.4 & 43.6 & 47.0 & 36.9 & 43.2 \\ %\hline
        10 & 64.1 & 65.8 & 62.7 & 67.5 & 72.8 & 72.7 & 44.2 & 47.1 & 37.1 & 43.3 \\ %\hline
        11 & 64.4 & 66.1 & 62.8 & 67.6 & 73.4 & 72.8 & 44.5 & 47.3 & 37.3 & 43.4 \\ %\hline
        12 & 64.6 & 66.0 & 62.5 & 67.4 & 74.0 & 73.1 & 44.7 & 47.4 & 37.5 & 43.7 \\ %\hline
        13 & 64.5 & 66.1 & 62.4 & 67.2 & 73.9 & 72.6 & 45.0 & 47.6 & 37.7 & 43.8 \\ %\hline
        14 & 64.9 & 66.4 & 62.5 & 67.3 & 74.3 & 73.4 & 45.2 & 47.6 & 37.8 & 43.9 \\ %\hline
        15 & 65.2 & 66.5 & 62.6 & 67.3 & 74.7 & 74.2 & 45.4 & 47.7 & 37.6 & 43.9 \\ %\hline
        16 & 65.6 & 66.6 & 62.7 & 67.4 & 75.2 & 74.0 & 45.4 & 48.2 & 37.8 & 44.1 \\ %\hline
        17 & 66.0 & 66.5 & 62.6 & 67.4 & 75.3 & 74.0 & 45.4 & 48.4 & 37.9 & 44.2 \\ %\hline
        18 & 65.9 & 66.7 & 62.5 & 67.4 & 75.3 & 74.6 & 45.3 & 48.5 & 37.8 & 43.7 \\ %\hline
        19 & 66.2 & 67.0 & 62.7 & 67.6 & 75.5 & 74.6 & 45.3 & 48.7 & 37.7 & 43.6 \\ %\hline
        20 & 66.4 & 67.1 & 62.7 & 67.7 & 74.9 & 74.5 & 45.3 & 48.8 & 37.9 & 43.7 \\ %\hline
        21 & 66.6 & 67.3 & 62.7 & 67.6 & 75.2 & 74.5 & 45.5 & 49.1 & 38.1 & 43.8 \\ %\hline
        22 & 67.1 & 68.1 & 62.7 & 67.7 & 75.1 & 74.5 & 45.6 & 49.2 & 38.2 & 43.7 \\ %\hline
        23 & 67.2 & 68.3 & 62.8 & 67.8 & 75.3 & 75.0 & 45.7 & 49.2 & 37.9 & 43.8 \\ %\hline
        24 & 67.2 & 68.3 & 62.9 & 67.9 & 75.4 & 75.2 & 45.8 & 49.3 & 37.9 & 43.8 \\ %\hline
\bottomrule  
\end{tabular}
\caption{Results ($\%$) on \textbf{validation set} for classification tasks in terms of MAP and F1 scores, using mention vectors computed by averaging first $\ell$ layers, and extracted from BERT.}
\label{klayer_svm_linear_bert_dev}
\end{table*}

% bert + test
\begin{table*}[p]
\centering
\footnotesize
\begin{tabular}{
@{}c
cc
cc
cc
cc
cc
}
\toprule  
\multirow{2}{0.3in}{\textbf{Layer}} 
& \multicolumn{2}{c}{\textbf{X-McRae}}
& \multicolumn{2}{c}{\textbf{CSLB}}
& \multicolumn{2}{c}{\textbf{Morrow}}
& \multicolumn{2}{c}{\textbf{WordNet}}
& \multicolumn{2}{c}{\textbf{BabelNet}}\\
\cmidrule(lr){2-3}
\cmidrule(lr){4-5}
\cmidrule(lr){6-7}
\cmidrule(lr){8-9}
\cmidrule(lr){10-11}
& MAP & F1        
& MAP & F1 
& MAP & F1 
& MAP & F1 
& MAP & F1\\
 \midrule
        1 & 64.5 & 55.7 & 58.6 & 42.8 & 62.1 & 50.1 & 33.7 & 36.3 & 33.8 & 31.6 \\ 
        2 & 64.3 & 54.9 & 58.9 & 44.2 & 62.4 & 50.8 & 34.9 & 37.2 & 35.0 & 34.7 \\ 
        3 & 65.1 & 57.1 & 59.3 & 44.7 & 65.5 & 51.2 & 36.2 & 38.6 & 35.2 & 34.9 \\ 
        4 & 66.2 & 57.2 & 59.7 & 44.4 & 66.0 & 53.9 & 37.0 & 38.5 & 36.0 & 35.3 \\ 
        5 & 67.2 & 57.8 & 60.2 & 45.4 & 67.7 & 56.9 & 38.9 & 39.4 & 37.4 & 35.1 \\ 
        6 & 68.2 & 58.0 & 61.5 & 45.8 & 69.8 & 58.2 & 40.1 & 40.3 & 38.8 & 35.4 \\ 
        7 & 68.9 & 60.2 & 62.4 & 47.0 & 72.4 & 61.0 & 41.0 & 41.8 & 39.5 & 36.6 \\ 
        8 & 69.2 & 59.4 & 62.7 & 47.9 & 74.0 & 60.3 & 42.0 & 42.2 & 40.4 & 37.7 \\ 
        9 & 69.4 & 60.2 & 62.8 & 47.9 & 74.8 & 62.8 & 42.8 & 42.6 & 41.3 & 40.3 \\ 
        10 & 70.0 & 59.1 & 62.9 & 48.4 & 75.3 & 62.7 & 43.1 & 42.7 & 41.4 & 39.8 \\ 
        11 & 69.9 & 58.4 & 62.9 & 48.3 & 75.5 & 63.7 & 43.6 & 43.1 & 41.6 & 40.3 \\ 
        12 & 70.1 & 58.3 & 62.9 & 48.7 & 75.8 & 64.2 & 44.0 & 43.1 & 41.6 & 39.7 \\ 
        13 & 70.1 & 58.7 & 62.9 & 48.7 & 75.7 & 62.5 & 44.4 & 42.5 & 41.9 & 39.8 \\ 
        14 & 70.2 & 59.3 & 62.9 & 49.3 & 75.7 & 61.6 & 44.7 & 44.0 & 41.6 & 40.5 \\ 
        15 & 70.2 & 59.7 & 63.2 & 48.9 & 75.6 & 61.6 & 45.0 & 44.0 & 41.8 & 40.4 \\ 
        16 & 70.4 & 60.0 & 63.1 & 49.5 & 75.9 & 63.8 & 45.5 & 43.5 & 41.8 & 40.5 \\ 
        17 & 70.4 & 59.4 & 63.2 & 49.5 & 75.9 & 63.1 & 45.8 & 42.8 & 41.9 & 40.1 \\ 
        18 & 70.1 & 59.6 & 63.2 & 49.1 & 75.7 & 62.7 & 45.9 & 43.0 & 42.2 & 40.8 \\ 
        19 & 69.9 & 59.7 & 63.1 & 49.6 & 75.9 & 62.5 & 46.0 & 43.0 & 42.2 & 40.5 \\ 
        20 & 70.1 & 60.3 & 63.2 & 49.4 & 75.4 & 62.2 & 46.2 & 44.1 & 42.3 & 40.2 \\ 
        21 & 70.2 & 60.5 & 63.4 & 49.1 & 75.1 & 62.4 & 46.2 & 43.9 & 42.6 & 40.6 \\ 
        22 & 70.4 & 60.5 & 63.3 & 49.4 & 75.3 & 63.0 & 46.4 & 44.7 & 42.6 & 40.5 \\
        23 & 70.5 & 60.7 & 63.3 & 49.1 & 75.8 & 63.4 & 46.5 & 44.9 & 42.3 & 40.9 \\ 
        24 & 70.6 & 60.9 & 63.3 & 48.8 & 76.4 & 62.9 & 46.6 & 44.8 & 42.2 & 41.0 \\
\bottomrule  
\end{tabular}
\caption{Results ($\%$) on \textbf{test set} for classification tasks in terms of MAP and F1 scores, using mention vectors computed by averaging first $\ell$ layers, and extracted from BERT.}
\label{klayer_svm_linear_bert_test}
\end{table*}

% roberta + dev
\begin{table*}[p]
\centering
\footnotesize
\begin{tabular}{
@{}c
cc
cc
cc
cc
cc
}
\toprule  
\multirow{2}{0.3in}{\textbf{Layer}} 
& \multicolumn{2}{c}{\textbf{X-McRae}}
& \multicolumn{2}{c}{\textbf{CSLB}}
& \multicolumn{2}{c}{\textbf{Morrow}}
& \multicolumn{2}{c}{\textbf{WordNet}}
& \multicolumn{2}{c}{\textbf{BabelNet}}\\
\cmidrule(lr){2-3}
\cmidrule(lr){4-5}
\cmidrule(lr){6-7}
\cmidrule(lr){8-9}
\cmidrule(lr){10-11}
& MAP & F1        
& MAP & F1 
& MAP & F1 
& MAP & F1 
& MAP & F1\\
 \midrule
        1 & 37.2 & 45.7 & 36.8 & 46.9 & 43.2 & 51.1 & 39.8 & 45.2 & 35.0 & 41.4 \\ %\hline
        2 & 37.9 & 45.9 & 38.0 & 47.9 & 46.7 & 53.1 & 40.6 & 45.9 & 35.6 & 41.0 \\ %\hline
        3 & 40.1 & 47.1 & 39.0 & 48.8 & 50.8 & 54.4 & 42.0 & 46.1 & 35.9 & 41.7 \\ %\hline
        4 & 43.0 & 49.6 & 40.9 & 50.2 & 53.9 & 56.2 & 43.1 & 47.0 & 36.0 & 42.6 \\ %\hline
        5 & 46.8 & 51.9 & 43.4 & 52.2 & 57.0 & 59.9 & 44.5 & 48.3 & 36.8 & 43.7 \\ %\hline
        6 & 49.7 & 54.0 & 46.0 & 54.2 & 60.8 & 62.2 & 45.9 & 49.5 & 37.6 & 44.3 \\ %\hline
        7 & 52.5 & 55.9 & 48.2 & 55.8 & 63.1 & 64.4 & 46.5 & 50.5 & 38.7 & 45.0 \\ %\hline
        8 & 54.9 & 57.2 & 50.4 & 57.5 & 65.6 & 66.0 & 47.1 & 51.1 & 39.5 & 45.6 \\ %\hline
        9 & 57.1 & 59.6 & 52.3 & 58.9 & 67.5 & 66.8 & 47.4 & 51.3 & 39.9 & 46.3 \\ %\hline
        10 & 58.4 & 60.5 & 53.4 & 60.1 & 68.7 & 68.2 & 47.9 & 51.7 & 40.3 & 46.5 \\ %\hline
        11 & 59.7 & 61.5 & 54.4 & 61.0 & 69.3 & 68.8 & 48.2 & 52.1 & 40.6 & 46.9 \\ %\hline
        12 & 60.3 & 61.9 & 55.2 & 61.7 & 70.0 & 69.7 & 48.7 & 52.6 & 40.8 & 47.0 \\ %\hline
        13 & 60.7 & 62.2 & 55.9 & 62.4 & 70.1 & 70.3 & 48.9 & 52.8 & 41.3 & 47.2 \\ %\hline
        14 & 61.5 & 62.6 & 56.7 & 62.8 & 70.7 & 70.3 & 49.1 & 52.8 & 41.3 & 47.4 \\ %\hline
        15 & 61.4 & 62.9 & 57.1 & 63.2 & 72.5 & 70.8 & 49.2 & 53.0 & 41.6 & 47.4 \\ %\hline
        16 & 61.7 & 63.4 & 57.3 & 63.5 & 71.8 & 70.1 & 49.4 & 52.9 & 41.6 & 47.6 \\ %\hline
        17 & 62.0 & 63.8 & 57.6 & 63.7 & 72.1 & 70.0 & 49.5 & 53.1 & 41.8 & 47.6 \\ %\hline
        18 & 62.2 & 64.3 & 58.0 & 63.9 & 72.4 & 70.9 & 49.6 & 53.1 & 41.6 & 47.6 \\ %\hline
        19 & 62.7 & 64.9 & 58.2 & 64.1 & 72.3 & 70.9 & 49.6 & 53.0 & 41.6 & 47.7 \\ %\hline
        20 & 63.1 & 65.4 & 58.0 & 64.2 & 72.4 & 70.9 & 49.6 & 52.9 & 41.6 & 47.7 \\ %\hline
        21 & 63.3 & 65.7 & 58.2 & 64.4 & 72.4 & 71.1 & 49.6 & 52.8 & 41.5 & 47.8 \\ %\hline
        22 & 63.5 & 66.0 & 58.3 & 64.5 & 72.7 & 70.7 & 49.6 & 52.7 & 41.4 & 47.9 \\ %\hline
        23 & 63.6 & 66.0 & 58.4 & 64.5 & 73.7 & 71.6 & 49.7 & 52.9 & 41.4 & 47.7 \\ %\hline
        24 & 63.5 & 66.0 & 58.5 & 64.5 & 73.8 & 71.9 & 49.7 & 53.0 & 41.4 & 47.7 \\ %\hline
\bottomrule  
\end{tabular}
\caption{Results ($\%$) on \textbf{validation set} for classification tasks in terms of MAP and F1 scores, using mention vectors computed by averaging first $\ell$ layers, and extracted from RoBERTa.}
\label{klayer_svm_linear_roberta_dev}
\end{table*}

% roberta + test
\begin{table*}[p]
\centering
\footnotesize
\begin{tabular}{
@{}c
cc
cc
cc
cc
cc
}
\toprule  
\multirow{2}{0.3in}{\textbf{Layer}} 
& \multicolumn{2}{c}{\textbf{X-McRae}}
& \multicolumn{2}{c}{\textbf{CSLB}}
& \multicolumn{2}{c}{\textbf{Morrow}}
& \multicolumn{2}{c}{\textbf{WordNet}}
& \multicolumn{2}{c}{\textbf{BabelNet}}\\
\cmidrule(lr){2-3}
\cmidrule(lr){4-5}
\cmidrule(lr){6-7}
\cmidrule(lr){8-9}
\cmidrule(lr){10-11}
& MAP & F1        
& MAP & F1 
& MAP & F1 
& MAP & F1 
& MAP & F1\\
 \midrule
        1 & 37.2 & 32.4 & 37.8 & 27.7 & 48.3 & 43.9 & 41.4 & 41.3 & 33.2 & 30.5 \\ %\hline
        2 & 39.0 & 34.8 & 38.4 & 28.2 & 49.1 & 43.7 & 41.3 & 42.1 & 34.3 & 33.0 \\ %\hline
        3 & 41.0 & 37.3 & 40.2 & 29.3 & 51.7 & 40.9 & 41.6 & 41.7 & 35.0 & 35.0 \\ %\hline
        4 & 43.8 & 38.9 & 41.6 & 29.8 & 56.3 & 46.5 & 42.8 & 43.0 & 36.3 & 34.6 \\ %\hline
        5 & 47.6 & 43.5 & 44.4 & 32.3 & 64.1 & 51.4 & 45.1 & 44.5 & 37.4 & 33.3 \\ %\hline
        6 & 50.6 & 45.4 & 47.9 & 34.2 & 67.5 & 56.1 & 46.9 & 44.0 & 38.8 & 35.6 \\ %\hline
        7 & 53.4 & 46.9 & 50.1 & 35.7 & 72.5 & 59.0 & 48.6 & 46.8 & 40.4 & 36.0 \\ %\hline
        8 & 55.4 & 47.3 & 52.2 & 37.0 & 74.3 & 58.8 & 49.6 & 47.8 & 41.2 & 38.0 \\ %\hline
        9 & 57.1 & 50.2 & 53.6 & 39.2 & 75.6 & 61.1 & 50.1 & 47.9 & 41.9 & 38.6 \\ %\hline
        10 & 58.2 & 51.5 & 55.0 & 40.3 & 76.2 & 61.5 & 50.7 & 49.2 & 42.4 & 39.5 \\ %\hline
        11 & 59.2 & 52.6 & 55.9 & 40.8 & 77.1 & 60.5 & 51.1 & 49.1 & 42.6 & 39.9 \\ %\hline
        12 & 60.1 & 51.9 & 56.5 & 41.2 & 77.4 & 61.1 & 51.2 & 49.5 & 42.8 & 38.7 \\ %\hline
        13 & 60.8 & 52.4 & 57.0 & 41.4 & 77.5 & 63.8 & 51.5 & 50.3 & 43.1 & 39.6 \\ %\hline
        14 & 61.4 & 53.6 & 57.5 & 42.1 & 77.7 & 64.5 & 51.6 & 49.7 & 43.0 & 39.5 \\ %\hline
        15 & 61.8 & 53.5 & 57.8 & 42.1 & 77.6 & 65.0 & 51.1 & 49.6 & 43.1 & 39.7 \\ %\hline
        16 & 61.9 & 54.3 & 58.1 & 42.0 & 77.6 & 64.9 & 51.6 & 49.8 & 43.3 & 39.2 \\ %\hline
        17 & 62.3 & 54.0 & 58.2 & 42.5 & 77.8 & 66.7 & 51.5 & 50.0 & 43.1 & 39.2 \\ %\hline
        18 & 62.8 & 54.0 & 58.3 & 42.4 & 77.7 & 67.2 & 51.6 & 50.2 & 42.9 & 38.9 \\ %\hline
        19 & 63.0 & 53.0 & 58.5 & 42.5 & 77.6 & 68.5 & 51.7 & 50.1 & 42.9 & 40.1 \\ %\hline
        20 & 63.0 & 53.2 & 58.7 & 42.5 & 77.9 & 67.8 & 51.6 & 49.4 & 43.1 & 40.0 \\ %\hline
        21 & 63.2 & 53.8 & 58.7 & 43.1 & 78.0 & 67.2 & 51.6 & 49.6 & 43.2 & 39.3 \\ %\hline
        22 & 63.4 & 53.3 & 58.7 & 43.5 & 77.9 & 67.4 & 51.3 & 49.7 & 43.3 & 39.3 \\ %\hline
        23 & 63.9 & 53.6 & 59.0 & 43.9 & 77.9 & 67.5 & 51.3 & 50.2 & 43.4 & 39.0 \\ %\hline
        24 & 63.9 & 53.5 & 58.9 & 44.2 & 77.9 & 67.5 & 51.3 & 49.8 & 43.2 & 38.7 \\ %\hline
\bottomrule  
\end{tabular}
\caption{Results ($\%$) on \textbf{test set} for classification tasks in terms of MAP and F1 scores, using mention vectors computed by averaging first $\ell$ layers, and extracted from RoBERTa.}
\label{klayer_svm_linear_roberta_test}
\end{table*}

%\bibliographystyle{named}
%\bibliography{ijcai21}
%\end{document}

\bibliographystyle{named}
\bibliography{ijcai21}

\begin{thebibliography}{}

\bibitem[\protect\citeauthoryear{Agirre \bgroup \em et al.\egroup
  }{2009}]{Agirreetal:09}
Eneko Agirre, Enrique Alfonseca, Keith Hall, Jana Kravalova, Marius Pa\c{s}ca,
  and Aitor Soroa.
\newblock A study on similarity and relatedness using distributional and
  {W}ord{N}et-based approaches.
\newblock In {\em Proceedings of NAACL}, pages 19--27, 2009.

\bibitem[\protect\citeauthoryear{Amrami and Goldberg}{2019}]{amrami2019towards}
Asaf Amrami and Yoav Goldberg.
\newblock Towards better substitution-based word sense induction.
\newblock {\em arXiv:1905.12598}, 2019.

\bibitem[\protect\citeauthoryear{Bommasani \bgroup \em et al.\egroup
  }{2020}]{DBLP:conf/acl/BommasaniDC20}
Rishi Bommasani, Kelly Davis, and Claire Cardie.
\newblock Interpreting pretrained contextualized representations via reductions
  to static embeddings.
\newblock In {\em Proceedings ACL}, pages 4758--4781, 2020.

\bibitem[\protect\citeauthoryear{Bruni \bgroup \em et al.\egroup
  }{2014}]{bruni2014multimodal}
Elia Bruni, Nam-Khanh Tran, and Marco Baroni.
\newblock Multimodal distributional semantics.
\newblock {\em J. Artif. Intell. Res.}, 49(1-47), 2014.

\bibitem[\protect\citeauthoryear{Camacho-Collados and
  Navigli}{2017}]{camacho2017babeldomains}
Jose Camacho-Collados and Roberto Navigli.
\newblock {BabelDomains}: Large-scale domain labeling of lexical resources.
\newblock In {\em Proc.\ EACL}, pages 223--228, 2017.

\bibitem[\protect\citeauthoryear{Camacho-Collados \bgroup \em et al.\egroup
  }{2017}]{camacho-collados-etal-2017-semeval}
Jose Camacho-Collados, Mohammad~Taher Pilehvar, Nigel Collier, and Roberto
  Navigli.
\newblock {S}em{E}val-2017 task 2: Multilingual and cross-lingual semantic word
  similarity.
\newblock In {\em Proc.\ {S}em{E}val}, pages 15--26, 2017.

\bibitem[\protect\citeauthoryear{Ciaramita and
  Johnson}{2003}]{ciaramita-johnson-2003-supersense}
Massimiliano Ciaramita and Mark Johnson.
\newblock Supersense tagging of unknown nouns in {W}ord{N}et.
\newblock In {\em Proc.\ EMNLP}, pages 168--175, 2003.

\bibitem[\protect\citeauthoryear{Das \bgroup \em et al.\egroup
  }{2015}]{das2015gaussian}
Rajarshi Das, Manzil Zaheer, and Chris Dyer.
\newblock Gaussian {LDA} for topic models with word embeddings.
\newblock In {\em Proc.\ ACL}, pages 795--804, 2015.

\bibitem[\protect\citeauthoryear{Devlin \bgroup \em et al.\egroup
  }{2019}]{DBLP:conf/naacl/DevlinCLT19}
Jacob Devlin, Ming{-}Wei Chang, Kenton Lee, and Kristina Toutanova.
\newblock {BERT:} pre-training of deep bidirectional transformers for language
  understanding.
\newblock In {\em Proc.\ NAACL-HLT}, 2019.

\bibitem[\protect\citeauthoryear{Ethayarajh}{2019}]{DBLP:conf/emnlp/Ethayarajh19}
Kawin Ethayarajh.
\newblock How contextual are contextualized word representations? comparing the
  geometry of {BERT}, {ELMo}, and {GPT-2} embeddings.
\newblock In {\em Proc.\ EMNLP}, pages 55--65, 2019.

\bibitem[\protect\citeauthoryear{Finkelstein \bgroup \em et al.\egroup
  }{2002}]{Levetal:2002}
Lev Finkelstein, Gabrilovich Evgenly, Matias Yossi, Rivlin Ehud, Solan Zach,
  Wolfman Gadi, and Ruppin Eytan.
\newblock Placing search in context: The concept revisited.
\newblock {\em ACM Transactions on Information Systems}, 20(1):116--131, 2002.

\bibitem[\protect\citeauthoryear{Forbes \bgroup \em et al.\egroup
  }{2019}]{forbes2019neural}
Maxwell Forbes, Ari Holtzman, and Yejin Choi.
\newblock Do neural language representations learn physical commonsense?
\newblock {\em Proc.\ CogSci}, 2019.

\bibitem[\protect\citeauthoryear{Hill \bgroup \em et al.\egroup
  }{2015}]{hill-etal-2015-simlex}
Felix Hill, Roi Reichart, and Anna Korhonen.
\newblock {S}im{L}ex-999: Evaluating semantic models with (genuine) similarity
  estimation.
\newblock {\em Computational Linguistics}, 41(4):665--695, 2015.

\bibitem[\protect\citeauthoryear{Kolyvakis \bgroup \em et al.\egroup
  }{2018}]{kolyvakis2018deepalignment}
Prodromos Kolyvakis, Alexandros Kalousis, and Dimitris Kiritsis.
\newblock Deepalignment: Unsupervised ontology matching with refined word
  vectors.
\newblock In {\em Proc.\ NAACL-HLT}, pages 787--798, 2018.

\bibitem[\protect\citeauthoryear{Li \bgroup \em et al.\egroup
  }{2019}]{li2019ontology}
Na~Li, Zied Bouraoui, and Steven Schockaert.
\newblock Ontology completion using graph convolutional networks.
\newblock In {\em Proc.\ ISWC}, pages 435--452, 2019.

\bibitem[\protect\citeauthoryear{Liu \bgroup \em et al.\egroup
  }{2019}]{DBLP:journals/corr/abs-1907-11692}
Yinhan Liu, Myle Ott, Naman Goyal, Jingfei Du, Mandar Joshi, Danqi Chen, Omer
  Levy, Mike Lewis, Luke Zettlemoyer, and Veselin Stoyanov.
\newblock {RoBERTa}: {A} robustly optimized {BERT} pretraining approach.
\newblock {\em CoRR}, abs/1907.11692, 2019.

\bibitem[\protect\citeauthoryear{Ma \bgroup \em et al.\egroup
  }{2016}]{ma2016label}
Yukun Ma, Erik Cambria, and Sa~Gao.
\newblock Label embedding for zero-shot fine-grained named entity typing.
\newblock In {\em Proc.\ COLING}, pages 171--180, 2016.

\bibitem[\protect\citeauthoryear{{McRae et al.}}{2005}]{mcrae2005semantic}
Ken {McRae et al.}
\newblock Semantic feature production norms for a large set of living and
  nonliving things.
\newblock {\em Behavior research methods}, 37:547--559, 2005.

\bibitem[\protect\citeauthoryear{Mickus \bgroup \em et al.\egroup
  }{2019}]{mickus2019you}
Timothee Mickus, Denis Paperno, Mathieu Constant, and Kees van Deemeter.
\newblock What do you mean, {BERT}? assessing {BERT} as a distributional
  semantics model.
\newblock {\em arXiv:1911.05758}, 2019.

\bibitem[\protect\citeauthoryear{Mikolov \bgroup \em et al.\egroup
  }{2013}]{DBLP:journals/corr/abs-1301-3781}
Tomas Mikolov, Kai Chen, Greg Corrado, and Jeffrey Dean.
\newblock Efficient estimation of word representations in vector space.
\newblock In {\em Proc.\ ICLR}, 2013.

\bibitem[\protect\citeauthoryear{Morrow and
  Duffy}{2005}]{morrow2005representation}
Lorna~I Morrow and M~Frances Duffy.
\newblock The representation of ontological category concepts as affected by
  healthy aging: Normative data and theoretical implications.
\newblock {\em Behavior research methods}, 37(4):608--625, 2005.

\bibitem[\protect\citeauthoryear{Onal \bgroup \em et al.\egroup
  }{2018}]{onal2018neural}
Kezban~Dilek Onal, Ye~Zhang, Ismail~Sengor Altingovde, Md~Mustafizur Rahman,
  Pinar Karagoz, Alex Braylan, Brandon Dang, Heng-Lu Chang, Henna Kim, Quinten
  McNamara, et~al.
\newblock Neural information retrieval: At the end of the early years.
\newblock {\em Information Retrieval Journal}, 21(2-3):111--182, 2018.

\bibitem[\protect\citeauthoryear{Pennington \bgroup \em et al.\egroup
  }{2014}]{DBLP:conf/emnlp/PenningtonSM14}
Jeffrey Pennington, Richard Socher, and Christopher~D. Manning.
\newblock {GloVe}: Global vectors for word representation.
\newblock In {\em Proc.\ EMNLP}, pages 1532--1543, 2014.

\bibitem[\protect\citeauthoryear{Rubenstein and Goodenough}{1965}]{RG65:1965}
Herbert Rubenstein and John~B. Goodenough.
\newblock Contextual correlates of synonymy.
\newblock {\em Communications of the ACM}, 8(10):627--633, 1965.

\bibitem[\protect\citeauthoryear{Socher \bgroup \em et al.\egroup
  }{2013}]{socher2013zero}
Richard Socher, Milind Ganjoo, Christopher~D Manning, and Andrew Ng.
\newblock Zero-shot learning through cross-modal transfer.
\newblock In {\em Proc.\ NIPS}, pages 935--943, 2013.

\bibitem[\protect\citeauthoryear{Vimercati \bgroup \em et al.\egroup
  }{2019}]{vimercati2019mapping}
Manuel Vimercati, Federico Bianchi, Mauricio Soto, and Matteo Palmonari.
\newblock Mapping lexical knowledge to distributed models for ontology concept
  invention.
\newblock In {\em Proc.\ IA*AI}, pages 572--587, 2019.

\bibitem[\protect\citeauthoryear{Vulic \bgroup \em et al.\egroup
  }{2020}]{DBLP:conf/emnlp/VulicPLGK20}
Ivan Vulic, Edoardo~Maria Ponti, Robert Litschko, Goran Glavas, and Anna
  Korhonen.
\newblock Probing pretrained language models for lexical semantics.
\newblock In {\em Proceedings EMNLP}, pages 7222--7240, 2020.

\bibitem[\protect\citeauthoryear{Weir \bgroup \em et al.\egroup
  }{2020}]{weir2020existence}
Nathaniel Weir, Adam Poliak, and Benjamin Van~Durme.
\newblock On the existence of tacit assumptions in contextualized language
  models.
\newblock {\em arXiv:2004.04877}, 2020.

\end{thebibliography}
\end{document}